\documentclass[runningheads]{llncs}

\usepackage{eccv}

\usepackage{eccvabbrv}

\usepackage{graphicx}
\usepackage{booktabs}

\usepackage[accsupp]{axessibility}  

\usepackage{algorithm}
\usepackage{algorithmic}
\usepackage{color}
\usepackage{multirow}
\usepackage{threeparttable}

\usepackage{hyperref}

\usepackage{orcidlink}

\begin{document}

\setlength\textfloatsep{19pt}

\title{Spectral Prior for Reducing Exposure Bias in Diffusion Models}


\author{Yuya Kobayashi\inst{1} \and
Masato Ishii\inst{1} \and
Yuhta Takida\inst{1} \and
Takashi Shibuya\inst{1}\orcidlink{0000-0002-4277-0164} \and\\
Yuki Mitsufuji\inst{1,2}\orcidlink{0000-0002-6806-6140}}

\authorrunning{Y.~Kobayashi et al.}

\institute{Sony AI \and Sony Group Corporation}

\maketitle

\begin{abstract}

Diffusion models typically suffer from error accumulation during iterative sampling, commonly referred to as exposure bias. We reveal systematic frequency-dependent discrepancies between training and inference, which can be interpreted as frequency-dependent SNR error. Crucially, the direction of this mismatch varies across models and timesteps, indicating that fixed correction rules do not generalize. We propose Spectral Alignment (SPA), a lightweight, guidance-based method that calibrates the power spectrum of intermediate predictions to a pre-computed prior. Our approach consists of two stages: (1) offline fitting of a parametric spectrum model from training data, and (2) inference-time guidance via efficient FFT-based gradient computation. SPA introduces minimal computational overhead (3-4\%) and is complementary to Classifier-Free Guidance (CFG). We demonstrate consistent improvements across diverse architectures, from pixel-space models (DDPM, ADM) to latent diffusion models (SD2.0, SDXL) and flow-matching models (SD3.5, FLUX). Our implementation is available at \url{https://github.com/SonyResearch/SPA}.
\end{abstract}
    
\section{Introduction}
\label{sec:intro}

Diffusion models have become a dominant paradigm for image generation~\cite{ddpm2020, dhariwal2021diffusion, song2021denoising, rombach2022ldm, lipman2023flow}. Despite this rapid progress, a core challenge remains: during iterative sampling, small prediction errors can accumulate and lead to a distribution shift away from the target, a phenomenon often referred to as \emph{exposure bias}. Prior studies have argued that exposure bias in diffusion sampling manifests primarily as a mismatch between the noise levels seen during training and those encountered at test time. Accordingly, these methods aim to correct the effective noise level to match the predefined noise schedule, e.g., by predicted-noise rescaling~\cite{ning2024elucidating}, time-shifted sampling~\cite{li2024alleviating}, and wavelet-based frequency reweighting~\cite{yu2025wavelet}. However, these approaches either assume frequency-uniform errors or rely on hand-designed correction rules and carefully tuned schedules.

In this work, we analyze the power spectrum of intermediate predictions $\hat{x}_{0|t}$ during diffusion inference and reveal a systematic spectral mismatch between the forward process and inference trajectories (Figure~\ref{fig:spectral_mismatch}).
We interpret this discrepancy as frequency-dependent effective SNR errors that vary across models and timesteps, and show that correcting it leads to improved image quality. We discuss the details of this observation and its implications for method design in Section~\ref{sec:motivation}.

\begin{figure}[t]
    \centering
    \includegraphics[width=\linewidth]{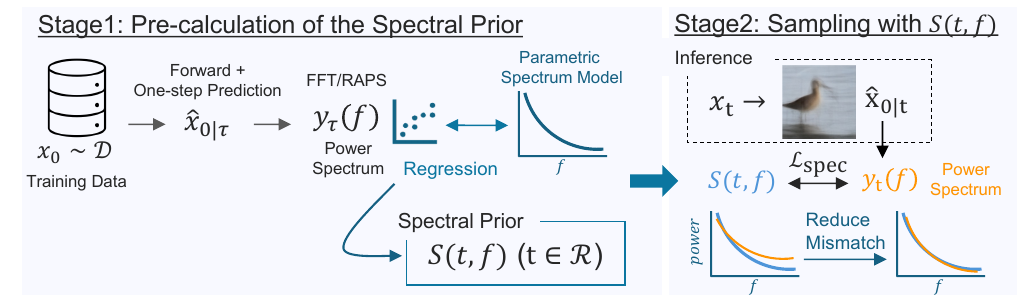}
    \caption{Overview of the proposed method. We pre-calculate the spectral prior in the first stage, and spectral mismatch is corrected using the prior in the second stage during inference. For latent diffusion models, apply the same procedure on latent $z$.}
    \label{fig:overview}
\end{figure}

Based on this observation, we propose \textbf{Spectral Alignment (SPA)}, a lightweight guidance-based calibration method applicable to a wide range of models. This method shifts the power spectra of intermediate predictions toward a precomputed timestep-dependent \textit{spectral prior}. The overview of the method is shown in Figure~\ref{fig:overview}. Our approach consists of two stages: (1) pre-calculation of the spectral prior and (2) sampling with guidance using spectral prior. In the first stage, a parametric spectrum model is fitted using training data. In the second stage, samples are steered toward the spectral prior via efficient FFT-based gradient computation.





The method can be applied to various models adaptively.
We conducted experiments across a wide range of text-to-image diffusion models, from early architectures (DDPM~\cite{ddpm2020}, ADM~\cite{dhariwal2021diffusion}) to LDMs (SD2.0~\cite{sd20}, SDXL~\cite{SDXL2024}) and state-of-the-art flow-matching models (SD3.5~\cite{sd35}, FLUX~\cite{flux}). We show that SPA outperforms other baseline methods, and consistently improves generation quality as measured by FID~\cite{TTUR} and reward models such as HPSv3~\cite{Ma_2025_ICCV}. In addition, SPA introduces minimal computational overhead (3--4\%) and can be easily integrated into existing generation pipelines.

Our main contributions are as follows:
\begin{itemize}
\item We reveal that even modern flow-matching models exhibit spectral mismatch, and propose a calibration method using a precomputed reference power spectrum. 
\item We demonstrate that correcting spectral mismatch with our approach improves generation quality with minimal computational overhead (3--4\%).
\item We show that SPA is applicable to a wide variety of models without modification, from DDPM to state-of-the-art flow-matching models.
\end{itemize}
\section{Motivation}
\label{sec:motivation}

In this section, we describe the key observations motivating the proposed method. We first introduce the notion of spectral mismatch from a frequency-dependent effective SNR error, and then present empirical observations that guided the design of our approach.

\subsection{Preliminaries}
\label{sec:preliminaries}

\subsubsection{Diffusion Models}
The forward process of diffusion models gradually adds Gaussian noise to data $x_0$ according to a predefined noise schedule. The latent variable at an intermediate timestep, denoted by $x_t$, can be obtained as:
\begin{equation}
x_t = \sqrt{\bar{\alpha}_t} x_0 + \sqrt{1-\bar{\alpha}_t} \epsilon, \quad \epsilon \sim \mathcal{N}(0, I),
\end{equation}
where $\bar{\alpha}_t$ determines the noise level at timestep $t$. The reverse process learns to denoise $x_t$ by predicting the noise $\epsilon_\theta(x_t, c, t)$ using a neural network, given condition $c$. The denoising step from $x_t$ to $x_{t-1}$ using DDIM~\cite{song2021denoising} is given by a deterministic update:
\begin{equation}
\label{eq:reverse}
x_{t-1} = \sqrt{\bar{\alpha}_{t-1}} \hat{x}_{0\mid t} + \sqrt{1-\bar{\alpha}_{t-1}} \cdot \epsilon_\theta(x_t, c, t),
\end{equation}
where $\hat{x}_{0\mid t}$ is the predicted clean data estimated from $x_t$ using Tweedie's formula:
\begin{equation}
\label{eq:tweedie}
\hat{x}_{0|t} = \frac{x_t - \sqrt{1-\bar{\alpha}_t} \cdot \epsilon_\theta(x_t, c, t)}{\sqrt{\bar{\alpha}_t}}.
\end{equation}


\subsubsection{Classifier-Free Guidance (CFG)}
For conditional generation, CFG~\cite{ho2021cfg} computes a guided noise prediction by extrapolating between the conditional and unconditional estimates:
\begin{equation}
\epsilon^{\text{CFG}}_\theta = \epsilon_\theta(x_t, t, \emptyset) + w \cdot \bigl( \epsilon_\theta(x_t, t, c) - \epsilon_\theta(x_t, t, \emptyset)\bigr),
\label{eq:cfg}
\end{equation}
where $w > 1$ is the guidance scale and $\emptyset$ denotes null conditioning. While CFG improves prompt adherence, high guidance scales often lead to oversaturation artifacts~\cite{wang2024analysis, sadat2025eliminating}.

\subsection{Spectral Mismatch}
\label{sec:spectral_mismatch}



\subsubsection{Frequency-Dependent SNR}


Exposure bias in diffusion models refers to the mismatch between the distribution of intermediate variables observed during training (\ie, forward process) and those during inference (\ie, reverse process).
Prior work suggests quantifying this mismatch through deviations from the predefined noise schedule, which is often referred to as an SNR error. Ning~\etal~\cite{ning2024elucidating} proposed rescaling the predicted noise to correct such errors. Furthermore, Yu~\etal~\cite{yu2025wavelet} showed that the mismatch differs between low- and high-frequency bands, indicating a frequency-dependent structure.
In this work, we go beyond these perspectives and examine how the mismatch varies across the entire frequency spectrum, timesteps, and model choices, and how it can be corrected.

To analyze this, we define the signal-to-noise ratio at frequency $\omega$ and timestep $t$ as:
\begin{equation}
\text{SNR}(\omega, t) = \frac{\bar{\alpha}_t\, |\tilde{x}_0(\omega)|^2}{(1-\bar{\alpha}_t)\, |\tilde{\epsilon}(\omega)|^2},
\label{eq:freq_snr}
\end{equation}
where $\tilde{x}_0(\omega) = \mathcal{F}[x_0]$ and $\tilde{\epsilon}(\omega) = \mathcal{F}[\epsilon]$ denote the Fourier coefficients at frequency $\omega$. Since Gaussian noise has a flat power spectrum (\ie, $\mathbb{E}[|\tilde{\epsilon}(\omega)|^2]$ is constant across frequencies), the frequency dependence of $\text{SNR}(\omega, t)$ is primarily determined by the signal power $|\tilde{x}_0(\omega)|^2$. Accordingly, rather than directly estimating an SNR error, we focus on the mismatch in signal strength (power spectrum) between training and inference in this study.

During training, ground-truth images $x_0$ are available, and their frequency characteristics can be computed directly. During inference, however, $x_0$ is unknown, preventing a direct comparison of $\text{SNR}(\omega, t)$ between the two settings. To address this, we use the estimated clean image $\hat{x}_{0|t}$ as a common proxy in both cases. Specifically, we compute $\hat{x}_{0|t}^{\text{train}}$ by applying a trained denoiser to $x_t$ sampled from the forward process, and $\hat{x}_{0|t}^{\text{infer}}$ from the reverse process at the corresponding timestep. By comparing the power spectra of these two estimates, we isolate the mismatch caused by the distributional gap between training-time and inference-time inputs, excluding the intrinsic high-frequency attenuation inherent to the posterior mean estimator. This allows us to quantify how the spectral mismatch varies across timesteps and models.

\subsubsection{Empirical Observation}

\begin{figure}[t]
    \centering
    \includegraphics[width=\linewidth]{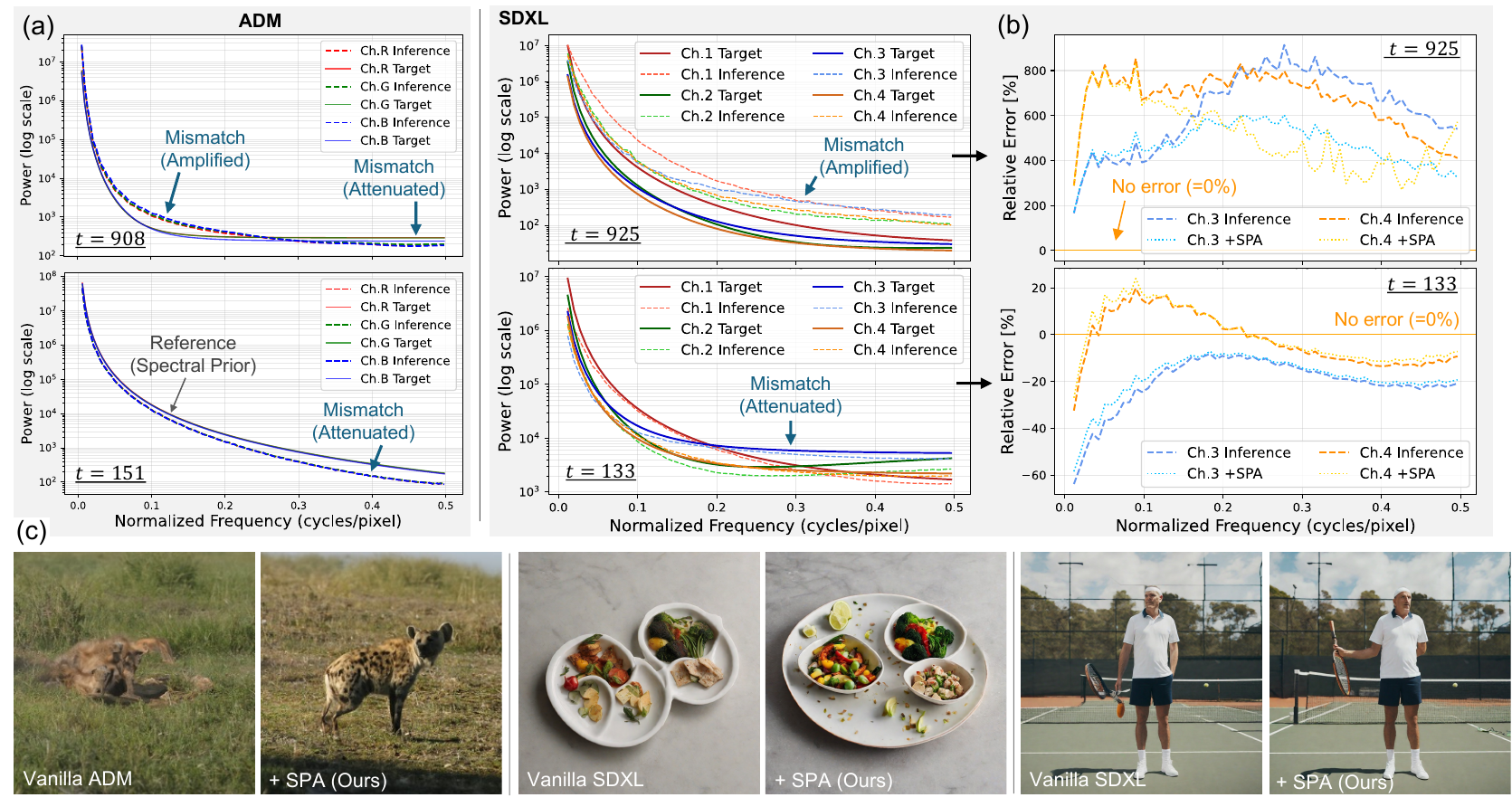}
    \caption{Visualization of spectral mismatch in ADM and SDXL. The channel-wise spectra are obtained from $\hat{x}_{0\mid t}$ (or $\hat{z}_{0\mid t}$ with SDXL) by averaging multiple samples. (a) Spectra from the target spectral prior (real lines) and from inference (broken lines). (b) Relative error for SDXL between the target and inference corresponding to the left figure. Results with SPA are also shown (dotted lines). They have smaller errors across frequencies (broken lines). Only two channels are visualized for visibility. (c) Examples of generated images used for these spectra. Each pair of images is sampled with the same initial noise and a random seed.}
    
    \label{fig:spectral_mismatch}
\end{figure}

Figure~\ref{fig:spectral_mismatch}(a) shows the averaged power spectra for each channel of $\hat{x}_{0|t}^\text{train}$ and $\hat{x}_{0|t}^\text{infer}$ obtained from ADM and SDXL. We observe systematic spectral mismatches across all models tested, but crucially, the direction and pattern of the mismatch vary across models and timesteps. For instance, ADM~\cite{dhariwal2021diffusion} exhibits high-frequency attenuation, while Stable Diffusion 2.0~\cite{sd20} shows low-frequency attenuation. More recent models such as SDXL~\cite{SDXL2024}, SD3.5~\cite{sd35}, and FLUX~\cite{flux} show complex, channel-wise, time-dependent patterns. More examples can be found in the Appendix.

\subsection{Motivation for the Proposed Method}
\label{sec:motivation_subsec}
A natural question is whether reducing the spectral mismatch directly improves sample quality, and if so, how to reduce it efficiently. In this paper, we propose a guidance-based method to reduce the spectral mismatch for two reasons. 
First, we aim to establish that spectral mismatch is a meaningful source of quality degradation and that correcting it yields measurable improvements. Second, we seek a practical, lightweight add-on applicable to existing pretrained models without retraining.

The model-dependent diversity shown in the observations above has two important implications.
First, fixed correction rules (e.g., ``boost high frequencies'') cannot generalize across architectures and denoising schedules.
Second, the mismatch carries spatial structure that cannot be resolved by scalar variance correction alone.
These observations motivate a data-driven approach that learns the appropriate target spectrum for each model.

\section{Method}
\label{sec:method}

\subsection{Overview}
\label{sec:method_overview}
In this section, we describe our guidance-based spectral mismatch correction method. Our approach consists of two stages.
In the first stage, we characterize the expected power spectrum of intermediate predictions $\hat{x}_{0|t}$ by fitting a parametric model using training data and a pretrained diffusion model. This is an offline process required only once per model.
In the second stage, during inference, we steer each denoising step toward the target spectrum via an efficient FFT-based gradient correction. We describe the details of the first stage in Section~\ref{sec:target_modeling} and those of the second stage in Section~\ref{sec:guidance}. Figure~\ref{fig:overview} shows an overview of our method.

\paragraph{Remark.}

This modeling process involves single-step model predictions (i.e., the forward process followed by one denoising step). While single-step prediction is not free of prediction error, it does not suffer from the error accumulation that arises during multi-step rollouts. 



\subsection{Target Power Spectrum Modeling}
\label{sec:target_modeling}
To define a correction target, we characterize the expected frequency properties of $\hat{x}_{0|t}$ by fitting a parametric spectrum model from training data.

\subsubsection{Power Spectrum Extraction}
For each channel of $\hat{x}_{0|t}$, we compute the 2D power spectrum $\mathbf{Y}_t = |\mathbf{F} \hat{x}_{0|t}|^2$, where $\mathbf{F}$ denotes the 2D DFT matrix, and $|\cdot|^2$ denotes the element-wise squared magnitude. We then reduce this to a 1D representation via radial averaging, commonly referred to as the Radially Averaged Power Spectrum (RAPS) \cite{RAPS}:
\begin{equation}
y_t(f) = \text{RadialAvg}(\mathbf{Y}_t),
\label{eq:radialavg}
\end{equation}
where $f \in (0, 0.5]$ is the normalized spatial frequency ($f=0.5$ is the Nyquist frequency). All operations are applied independently to each channel. 

\subsubsection{Parametric Spectrum Model}
\label{sec:spectrum_model}
It is well known that the power spectrum of natural images follows a power law~\cite{Field1987imagestat, VANDERSCHAAF1996}.
We confirmed that $\hat{x}_{0|t}$, which is basically a blurry image, also follows the power law with slightly different parameters from natural images.
We model the target spectrum as:
\begin{equation}
S(t, f) = p_t \cdot f^{-q_t} + r_t \cdot f + s_t,
\label{eq:power_law}
\end{equation}
where $p_t$, $q_t$, $r_t$, and $s_t$ are timestep-dependent parameters.
The first term captures the power-law decay characteristic of natural images.
The bias term $s_t$ and the linear term $r_t \cdot f$ represent deviations from the pure power law. For the linear term in particular, we set $r_t = 0$ by default and enable it only when it significantly improves the fit (e.g., SDXL, FLUX). All channels are modeled independently.

\paragraph{Model design.}
It is technically possible to use a 2D representation directly; however, we chose to use RAPS for the following reasons. (1) A 1D power spectrum is easier to visualize, which makes analysis, explanation, and hyperparameter tuning much simpler, and (2) the regression model becomes much simpler in the 1D case, which makes the fitting process more stable and more robust.

\subsubsection{Parameter Estimation Steps} \hfill
\paragraph{Step 1: Per-timestep fitting.}
For a discrete set of timesteps $\tau \in \{t_1, t_2, \ldots, t_K\}$, we collect reference power spectra via the following procedure:
\begin{enumerate}
\item Sample clean images $x_0^{(i)}$ from the training dataset and apply the forward process to obtain $x_\tau^{(i)} = \sqrt{\bar{\alpha}_\tau}\, x_0^{(i)} + \sigma_\tau\, \epsilon^{(i)}$.
\item Compute $\hat{x}_{0|\tau}^{(i)}$ using Eq.~\eqref{eq:tweedie} with the pretrained model.
\item Extract the RAPS $y_\tau^{(i)}(f)$ via Eq.~\eqref{eq:radialavg} and average across samples: $\bar{y}_\tau(f) = \frac{1}{N}\sum_{i=1}^{N} y_\tau^{(i)}(f)$.
\item Fit the parametric model (Eq.~\eqref{eq:power_law}) to $\bar{y}_\tau(f)$ via least-squares regression, obtaining $\{p_\tau, q_\tau, r_\tau, s_\tau\}$ for each channel and timestep independently.
\end{enumerate}

\noindent
\paragraph{Step 2: Temporal Interpolation.}
To obtain smooth parameter functions over all timesteps $t \in [0, T]$, we apply cubic spline interpolation to each parameter:
\begin{equation}
g(t) = \text{CubicSpline}(t;\{\tau_k, g_{\tau_k}\}),
\end{equation}
where $g \in \{p, q, r, s\}$. Since the number of evaluation steps during inference is not fixed, the resulting model $S(t, f)$ needs to capture the expected spectral characteristics of $\hat{x}_{0|t}$ at any timestep.

\subsection{Guidance-based Spectral Mismatch Calibration}
\label{sec:guidance}
Having established the target spectrum, we now describe how to steer the inference trajectory toward it.
We frame this as a guidance problem, adapting Diffusion Posterior Sampling (DPS)~\cite{chung2023diffusion} to enforce spectral alignment at each denoising step.
This inference-time approach serves as a practical add-on for existing pretrained models.

\paragraph{Guidance Loss.} \label{sec:loss}
At each reverse diffusion step, we compute $\hat{x}_{0|t}$ via Tweedie's formula (Eq.~\eqref{eq:tweedie}), extract its RAPS $y_t(f)$, and compare against the target spectrum $S(t, f)$.
The guidance loss is defined as:
\begin{equation}
\mathcal{L}(x_t, t) = \frac{1}{N}\sum_{c=1}^{N} \sum_{k=1}^{K} \bigl(\log_{10} y_t(f_k^c) - \log_{10} S(t, f_k^c)\bigr)^2,
\label{eq:guidance_loss}
\end{equation}
where the quadratic loss is summed over $K$ frequency bins and averaged over $N$ channels. $f_k \in (0, 0.5]$ represents the center frequency of the $k$-th bin.
We use an unweighted MSE loss in log-spectrum space for simplicity and stability, and leave frequency-dependent weighting for future work.

\paragraph{Asymmetric Penalty.}
\label{sec:penalty}
During the fitting process, we observed that the sample-wise median of the spectrum $\log_{10} y_\tau(f)$ from the forward process tends to be larger than the corresponding mean and exhibits positive skewness at each frequency $f$.
This indicates that $\log_{10} y_t(f)$--an observation during inference--falling too far below the target is undesirable, while exceeding the target is comparatively benign (see the Appendix for visualization).

This observation leads us to introduce an asymmetric penalty, which penalizes cases where the current spectrum falls below the target more strongly:
\begin{equation}
\phi_a(x) =
\begin{cases}
x & \text{if } x \geq 0, \\
a \cdot x & \text{if } x < 0,
\end{cases}
\label{eq:penalty}
\end{equation}
where $a > 1$ controls the penalty intensity.
The modified loss becomes:
\begin{equation}
\mathcal{L}_{\text{spec}}(x_t, t) = \frac{1}{N}\sum_{c=1}^{N}\sum_{k=1}^{K} \bigl(\phi_a(\log_{10} y_t(f_k^c) - \log_{10} S(t, f_k^c))\bigr)^2
\label{eq:guidance_loss_asym}
\end{equation}

\paragraph{Adaptation to LDMs.}
The method we described above can be applied to latent diffusion models without modification. Our assumption so far is that the power spectrum of the intermediate variable $x_t$ follows a power law. VAEs of latent diffusion models are known to preserve the spatial information from the pixel space \cite{rombach2022ldm}. 
In practice, the power spectrum of $z_t$ can be well fitted by our parametric power-law model. Visualization of the fitting can be found in the Appendix.

\paragraph{Overall Algorithm:}
Algorithm~\ref{alg:spectrum_guided_sampling} summarizes the complete sampling procedure with CFG.
Our spectral alignment is applied after the CFG combination step (line~4) and before the standard denoising update (line~9).
Thus, the method is generally compatible with most CFG variants~\cite{wang2024analysis, chung2025cfgplus, sadat2025eliminating, kwon2025tcfg} and other noise prediction modifications, since it operates on the combined prediction ${\epsilon^\text{CFG}}$ without any assumptions about how it was obtained.

\renewcommand{\algorithmicensure}{\textbf{Notation:}}

\begin{algorithm}[t]
\caption{Spectral Alignment (SPA) for Conditional Sampling}
\label{alg:spectrum_guided_sampling}
\begin{algorithmic}[1]
\REQUIRE Pretrained model $\epsilon_\theta$, target spectrum model $S(t, f)$, guidance strength $\eta$, penalty intensity $a$, text embedding $c$, null embedding $\emptyset$, and CFG scale $w$
\ENSURE $\mathrm{sg}[\cdot]$ denotes the stop-gradient operator.
\STATE Sample initial noise $x_T \sim \mathcal{N}(0, I)$
\FOR{$t = T, T-1, \ldots, 1$}
    \STATE $\epsilon_c, \epsilon_\emptyset \leftarrow \epsilon_\theta(x_t, t, c),\ \epsilon_\theta(x_t, t, \emptyset)$
    \STATE $\epsilon^{\mathrm{CFG}} \leftarrow \epsilon_\emptyset + w(\epsilon_c - \epsilon_\emptyset)$ \hfill $\triangleright$ \textit{Classifier-Free Guidance}
    \STATE $\hat{x}_{0|t} \leftarrow (x_t - \sigma_t\, \mathrm{sg}[\epsilon^{\mathrm{CFG}}]) / \sqrt{\bar{\alpha}_t}$ \hfill $\triangleright$ \textit{Tweedie's formula}
    \STATE $y_t \leftarrow \text{RadialAvg}(|\mathbf{F}\, \hat{x}_{0|t}|^2)$ \hfill $\triangleright$ \textit{Extract RAPS}
    \STATE $\mathcal{L}_{\text{spec}} \leftarrow \sum_{k=1}^{K} \bigl(\phi_a(\log_{10} y_t(f_k) - \log_{10} S(t, f_k))\bigr)^2$ \hfill $\triangleright$ \textit{Asymmetric Loss}
    \STATE $x_t \leftarrow x_t - \eta\, \nabla_{x_t} \mathcal{L}_{\text{spec}}$ \hfill $\triangleright$ \textit{Spectral Alignment}
    \STATE $x_{t-1} \leftarrow \text{Denoise}(x_t, \epsilon^{\text{CFG}}, t)$ \hfill $\triangleright$ \textit{Sampler step (DDIM, Euler, etc.)}
\ENDFOR
\RETURN $x_0$
\end{algorithmic}
\end{algorithm}

\paragraph{Computational Efficiency.}
The additional cost of our method is minimal.
Spectrum extraction via FFT has complexity $\mathcal{O}(D \log D)$ for $D$-dimensional data, and radial averaging is $\mathcal{O}(D)$. The gradient computation involves only elementwise operations and an inverse FFT for the backpropagation through the FFT, since it only involves the loss function, RAPS, and Tweedie's formula (lines 5--7 of Algorithm~\ref{alg:spectrum_guided_sampling}). Because the gradient from $\epsilon^\text{CFG}$ is not used, no additional neural network evaluations or backpropagation steps are required for guidance.
In practice, the per-step overhead is less than 5\% of the total sampling time (see Section~\ref{sec:overhead}).


\section{Experiments}
\label{sec:results}

\subsection{Experimental Setup}
\label{sec:setup}

\paragraph{Models.}
We evaluate our method across a wide range of diffusion architectures:
unconditional/class-conditional pixel-space models (DDPM~\cite{ddpm2020} on CelebA-HQ~\cite{karras2018progressive}, ADM on ImageNet 256$\times$256~\cite{imagenet}),
text-to-image latent diffusion models (Stable Diffusion 2.0, SDXL~\cite{SDXL2024}),
and flow-matching models (SD3.5 medium, FLUX.1\,[dev]). For SDXL, we used the first stage without the refiner. We used the DDPM sampler for ADM, which is the default in the official implementation, and the DDIM sampler for the other models. We used 50 denoising steps for DDPM, 100 steps for ADM, and 30 steps for the other models.

\paragraph{Pre-calculation of target spectra.}
We fit the parametric spectrum model (Section~\ref{sec:spectrum_model}) using 10K samples. The model's training dataset was used for DDPM and ADM (CelebA-HQ and ImageNet, respectively), and the LAION-Aesthetics V2 dataset~\cite{laion} was used for text-to-image models. The computational cost of this process is comparable to the scale of calculating metrics such as FID, and this process is only required once per model. 
We confirmed that the coefficient of determination ($\mathcal{R}^2$ score) exceeds 0.9 in most cases (see the Appendix for details).

\paragraph{Evaluation Metrics.}
\label{sec:eval_metrics}
For unconditional and class-conditional models (DDPM, ADM), we report FID and KID as primary metrics, which reliably measure distributional fidelity in unconditional settings. We used clean-fid~\cite{clean-fid} to compute these metrics. We additionally report Density and Coverage~\cite{prdc}, which are updated versions of Precision and Recall.
For text-to-image models (SD2.0, SDXL, SD3.5, FLUX), we adopt HPSv3 and ImageReward (reward-model-based metrics)~\cite{Ma_2025_ICCV, xu2023imagereward} as primary indicators, since FID is known to correlate negatively with human preference for these models, especially at high CFG scales~\cite{SDXL2024}.
We also report the CLIP score (ViT-G/14) as a measure of text alignment. For all evaluations, we use 5K prompts from the MS COCO validation set..

\paragraph{Baselines.}
\label{sec:baselines}
We compare against the following noise correction methods:
(1) $\epsilon$-rescaling~\cite{ning2024elucidating}, which rescales the predicted noise to match the expected norm;
(2) time-shift sampling~\cite{li2024alleviating}, which adjusts the noise schedule at inference time; and
(3) wavelet-based frequency regulation~\cite{yu2025wavelet}, which applies frequency-band reweighting.


Hyperparameters were determined as follows. For (1), we used 1.004, following their ImageNet experiment. For SD2.0 and SDXL, we evaluated a candidate set $\{1.003, 1.005, 1.007\}$ and selected $1.005$. For (2), the default values were used. The coefficient for (3) depends on the NFE and image resolution, which complicates the hyperparameter search. They reported the unimodality of FID with respect to the hyperparameters. Furthermore, since the qualitative effectiveness of this method becomes more prominent as the coefficient increases, we adopted the largest possible value within the range where no visual artifacts were observed. Specifically, we applied $(w^l, w^h)=(1.0005, 1.005)$ for ADM, $(1.003, 1.005)$ for SD2.0, and $(1.002, 1.001)$ for SDXL. Note that (1) has negligible computational overhead, (2) requires an additional timestep search (as detailed in Appendix F of their original work), and (3) involves a wavelet transform.

\paragraph{SPA Hyperparameters.}
The hyperparameters used for our proposed method are as follows. DDPM: $(\eta, a)=(0.01,1.0)$, ADM: $(0.05,2.0)$, SD2.0 and SDXL: $(0.2,5.0)$, SD3.5: $(0.75,3.0)$, FLUX.1\,[dev]: $(0.2,2.0)$ for CFG scale $w=2.5$ and $(0.2,1.0)$ for $w=3.5$. Here, $\eta$ is the guidance strength and $a$ is the asymmetric penalty intensity. To manage computational costs, we first performed a preliminary qualitative assessment to identify promising parameter ranges, followed by a focused quantitative evaluation to finalize these values. Although SPA introduces two additional hyperparameters, they typically require tuning only once per model architecture. Notably, models with similar designs (e.g., SD2.0 and SDXL) converged to identical optimal values despite being searched independently, suggesting the robustness and transferability of these parameters.


\subsection{Quantitative Results}
\label{sec:main_results}

\subsubsection{Unconditional and Class-Conditional Generation}
We evaluated our method on unconditional/class-conditional pixel-space models. Since all baselines are implemented for ADM, we conducted the comparison on ADM (ImageNet $256\times256$). The results are shown in Table~\ref{tab:uncond_results}. On ADM, our method consistently outperforms the other methods. Density/Coverage are also improved, even though SPA guides intermediate variables toward the average spectrum. This indicates that SPA has benefits in terms of generation diversity. The time-shift sampler did not perform well in our experiments. This is likely due to the difference in image resolution, as they only experimented with resolutions up to $128\times128$.

We also observe consistent improvement on DDPM. However, the performance gain is slightly smaller than that on ADM, and simple $\epsilon-$rescaling works slightly better.
This is likely because our method involves single-step model prediction in the target spectrum fitting stage. The target may be inaccurate when the single-step prediction itself is not accurate. We observed that the DDPM model did not perform well at predicting $\hat{x}_{0|t}$ even from ground-truth (forward process) training data $x_t$, which was not the case with the other models.

\begin{table}[t]
\centering
\caption{Results on unconditional and class-conditional generation.
}
\label{tab:uncond_results}
\begin{tabular}{llccccc}
\toprule
Model & Method & FID $\downarrow$ & KID$_{(\times1000)}$ $\downarrow$ & Density $\uparrow$ & Coverage $\uparrow$ \\
\midrule
\multirow{5}{*}{\shortstack{ADM \\ (ImageNet)}}
 & Vanilla              & 9.29  & 19.48 & 1.133 & 0.5939 \\
 & $\epsilon$-rescaling & 8.00  & 11.01 & 1.146 & 0.5976 \\
 & Time-shift   & 15.35 & 77.47 & 0.761 & 0.4321 \\
 & Wavelet reg.         & 8.35  & 12.80 & 1.122 & 0.5929 \\
 & \textbf{SPA (Ours)}  & \textbf{7.81} & \textbf{10.25} & \textbf{1.226} & \textbf{0.6184} \\
\midrule
\multirow{5}{*}{\shortstack{DDPM \\ (CelebA)}}
 & Vanilla              & 49.44 & 43.57 & 0.8186 & \textbf{0.0169} \\
 & $\epsilon$-rescaling & \textbf{48.53} & \textbf{40.98} & \textbf{0.8791} & \textbf{0.0169} \\
 & Time-shift           & 90.19 & 46.65 & 0.8440 & 0.0038 \\
 & Wavelet reg.         & 49.46 & 43.31 & 0.8182 & 0.0168 \\
 & \textbf{SPA (Ours)}  & 48.63 & 41.36 & 0.8344 & 0.0167 \\

\bottomrule
\end{tabular}
\end{table}

\begin{table}[t]
\centering
\caption{Quantitative results on text-to-image generation. We primarily evaluate image quality using HPSv3 and ImageReward, and additionally report CLIP Score (in gray) to indicate text alignment is preserved. $w$ denotes a CFG scale. $^\dagger$ The scores in parentheses uses true CFG (no distillation) with a ``de-distilled'' checkpoint~\cite{flux_dedistill}. See Section~\ref{sec:eval_metrics} for metric selection.}
\label{tab:t2i_results}
\begin{tabular}{llccc}
\toprule
Model & Method & HPSv3 $\uparrow$ & ImageReward $\uparrow$ & \textcolor{gray}{CLIP Score $\uparrow$} \\
\midrule
\multirow{5}{*}{\shortstack{SD2.0 \\ ($w$=7.5)}}
 & Vanilla              & 7.043 & 0.393 & \textcolor[gray]{0.5}{0.333} \\
 & $\epsilon$-rescaling & 7.186 & 0.415 & \textcolor[gray]{0.5}{0.333} \\
 & Time-shift           & 7.101 & 0.407 & \textcolor[gray]{0.5}{0.333} \\
 & Wavelet reg.         & 7.193 & \textbf{0.429} & \textcolor[gray]{0.5}{0.332} \\
 & \textbf{SPA (Ours)}  & \textbf{7.238} & 0.421 & \textcolor[gray]{0.5}{0.333} \\ 
\midrule
\multirow{5}{*}{\shortstack{SDXL \\ ($w$=7.5)}}
 & Vanilla              & 8.426 & 0.791 & \textcolor[gray]{0.5}{0.341} \\
 & $\epsilon$-rescaling & 8.528 & 0.800 & \textcolor[gray]{0.5}{0.340} \\
 & Time-shift           & 8.352 & 0.795 & \textcolor[gray]{0.5}{0.340} \\
 & Wavelet reg.         & 8.576 & 0.807 & \textcolor[gray]{0.5}{0.340} \\
 & \textbf{SPA (Ours)}  & \textbf{8.829} & \textbf{0.829} & \textcolor[gray]{0.5}{0.341} \\
\midrule
\multirow{2}{*}{\shortstack{SD3.5  \\ ($w$=5)}}
 & Vanilla              & 9.782 & 0.939 & \textcolor[gray]{0.5}{0.335} \\
 & \textbf{SPA (Ours)}  & \textbf{9.975} & \textbf{0.976} & \textcolor[gray]{0.5}{0.334} \\
\midrule
\multirow{2}{*}{\shortstack{FLUX.1 \\ ($w$=2.5)}}
 & Vanilla & 12.53 (11.55$^\dagger$) & 1.044 (0.989) & \textcolor[gray]{0.5}{0.328 (0.332)} \\
 & \textbf{SPA (Ours)} & \textbf{12.57} (\textbf{11.76}) & \textbf{1.054} (\textbf{1.005}) & \textcolor[gray]{0.5}{0.328 (0.332)} \\
\midrule
\multirow{2}{*}{\shortstack{FLUX.1 \\ ($w$=3.5)}}
 & Vanilla & 12.47 (11.94) & 1.070 (1.044) & \textcolor[gray]{0.5}{0.330 (0.334)} \\
 & \textbf{SPA (Ours)} & \textbf{12.48} (\textbf{12.02}) & 1.070 (\textbf{1.052}) & \textcolor[gray]{0.5}{0.330 (0.334)} \\
\bottomrule
\end{tabular}

\end{table}

\subsubsection{Text-to-Image Generation}
\label{sec:t2i_results}
Next, we evaluated text-to-image models. Although baseline methods were designed for ADM, we re-implemented them for SD2.0 and SDXL with minimal modifications. Table~\ref{tab:t2i_results} shows the quantitative results.

SPA also performs well across various T2I models. Note that scores from reward models behave more like an ordinal scale than an interval or ratio scale. With that in mind, we conclude that SPA performs best among these baselines, especially on SDXL (approximately a $4.5\%$ gain). CLIP Scores appear to be saturated, but SPA does not degrade text alignment.

SPA also works well with both flow-matching models. For FLUX.1, SPA achieves decent performance with true CFG (the scores in parentheses), which is the standard two-pass CFG without distillation, but the score improvement with guidance distillation (without parentheses) is relatively small. In the following section, we discuss the results for FLUX.1\,[dev] in detail.

\paragraph{Discussion on Guidance-Distilled Models.}
FLUX.1\,[dev] is a guidance-distilled model~\cite{meng2022on}, trained to internalize the effect of CFG. We observe that SPA provides statistically significant improvements at lower guidance scales ($w=2.5$, win-rate $53.3\pm1.5\%$, $p=2.3 \times 10^{-5}$), while the effect diminishes at higher scales ($w=3.5$). This suggests that guidance distillation may partially address spectral mismatch during training but does not fully eliminate it, particularly at lower guidance scales.

Notably, SPA shows stronger improvements on lower-quality samples. For images with HPSv3 scores in the bottom 20\%, the win-rate increases to $54.1 \pm 3.3\%$ ($p=0.02$) even at $w=3.5$. The average score gain for the bottom 5\% of samples is $0.70$ ($w=2.5$) and $0.19$ ($w=3.5$). This indicates that SPA serves as a useful refinement mechanism that selectively improves samples suffering from spectral distortion.




\subsubsection{Computational Overhead}
\label{sec:overhead}
We evaluated the computational overhead of SPA on SDXL. We compared the time consumed for (A) the standard denoising step (lines 3--4 in Algorithm~\ref{alg:spectrum_guided_sampling}) with (B) the additional steps required for SPA (lines 5--8). The average time for (A) was 2.47\,s and for (B) was 0.0952\,s; thus, the overhead is only $+3.86\%$. Since the FFT is the bottleneck, SPA works efficiently with latent diffusion models. 
We observed a similar or smaller overhead on FLUX.1\,[dev]. The relative overhead was only $+0.08\%$, since the FFT cost scales favorably with the latent spatial resolution. This confirms that the low overhead of SPA is not specific to SDXL.

\subsection{Qualitative Results}
\label{sec:qualitative}
\begin{figure}[t]
    \centering
    \includegraphics[width=\linewidth]{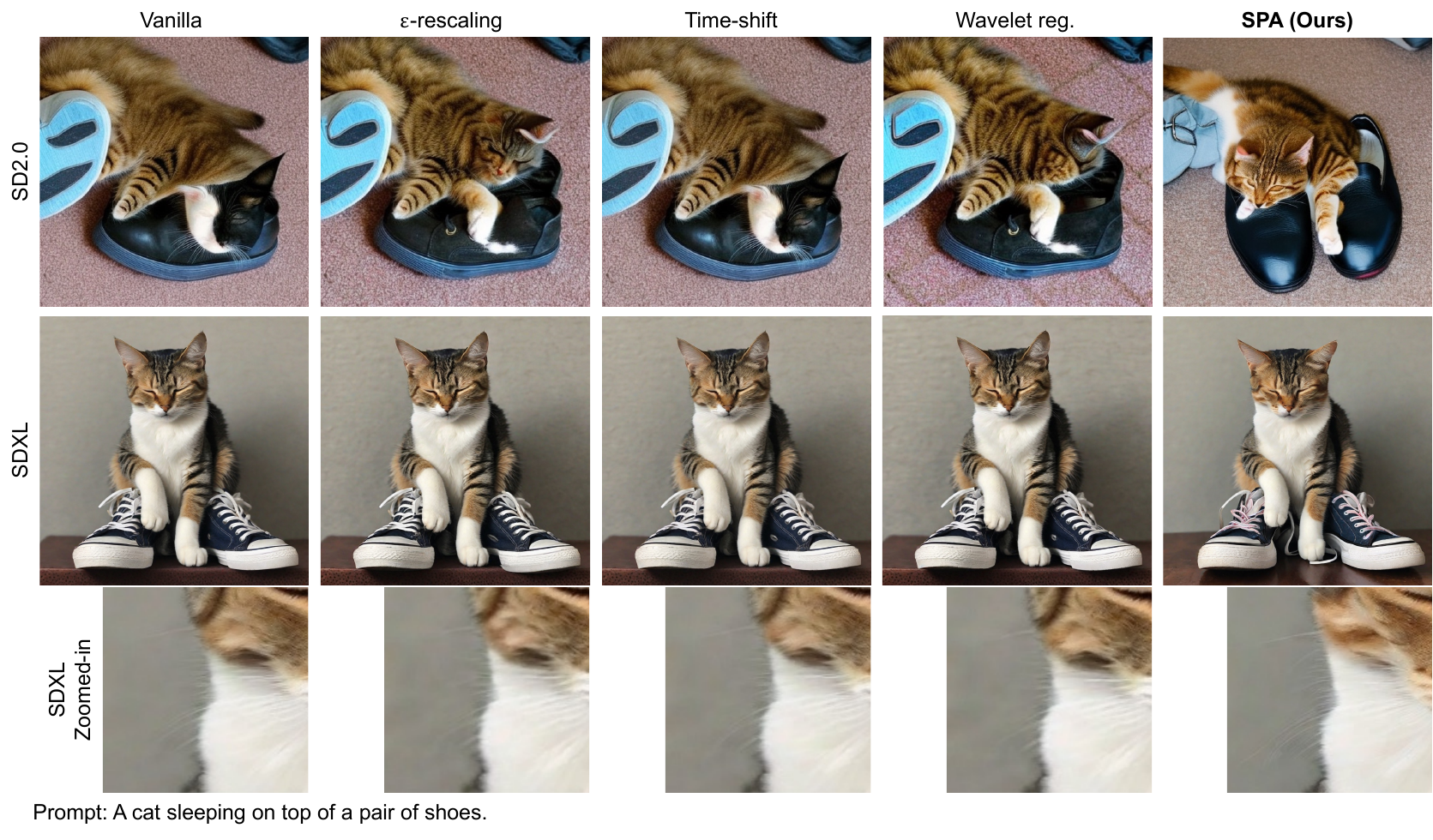}
    \caption{Qualitative comparison of generated images with baseline methods. The baseline methods are enumerated in Section~\ref{sec:baselines}. SPA successfully improved the sample quality and fixed the flawed object shapes in the SD2.0 sample. On the SDXL sample, SPA refined the edge definition (bottom row, zoomed), while baseline methods added frequency artifacts to the background.
    }
    \label{fig:qualitative_baselines}
\end{figure}

Figure~\ref{fig:qualitative_baselines} shows results for SD2.0 and SDXL with the baseline methods listed in Section~\ref{sec:baselines}. In our experiments, SPA tends to refine flawed object shapes, as shown in the SD2.0 example in this figure. When the vanilla inference is already decent, SPA keeps the content as is, while slightly enhancing the color and detail, as seen in the SDXL sample.

\begin{figure}[t]
    \centering
    \includegraphics[width=\linewidth]{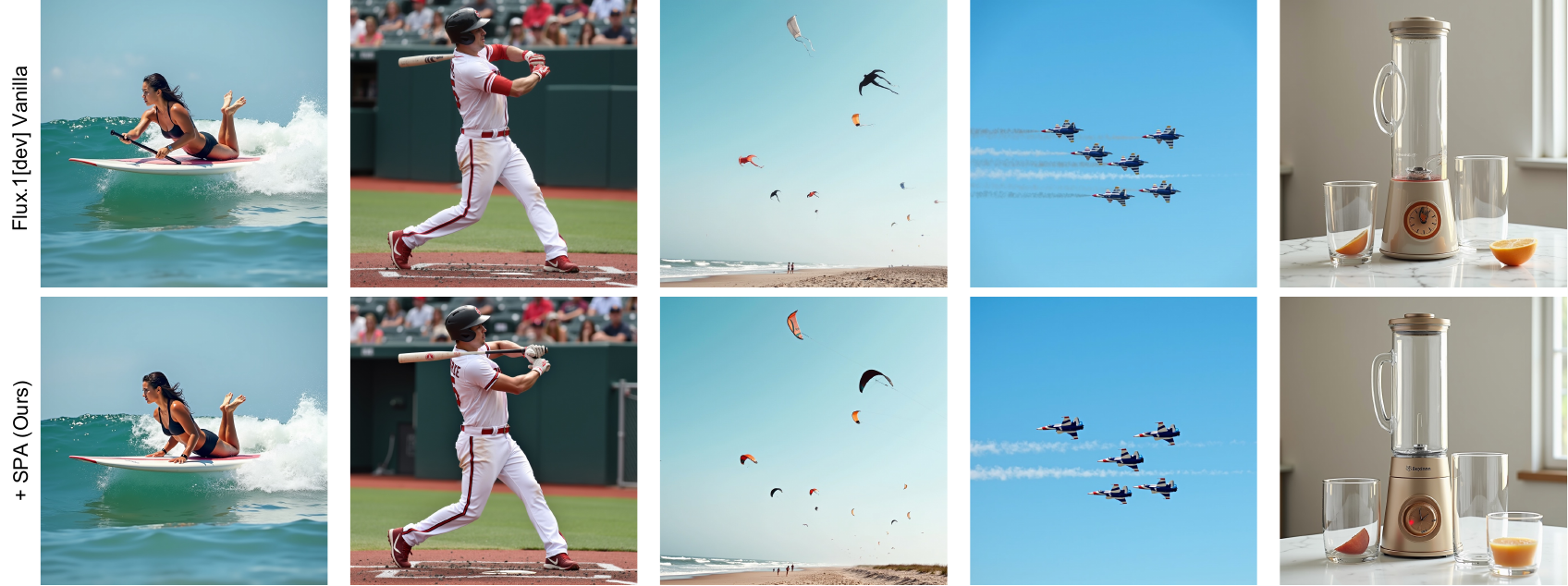}
    \caption{Qualitative results from FLUX.1\,[dev]. 
    Despite the strong baseline performance of FLUX.1\,[dev], it occasionally generates physically implausible structures (e.g., malformed objects, unnatural compositions). SPA selectively refines such failure cases without degrading already high-quality samples. This behavior is particularly valuable in practical deployment where consistent quality is required.
    }
    \label{fig:qualitative_flux}
\end{figure}

Figure~\ref{fig:qualitative_flux} shows results from FLUX.1\,[dev]. This model has strong baseline performance; however, the model still occasionally generates unnatural structures. In such cases, SPA can refine the images. This behavior is practically useful for making adjustments after fixing the layout (via seed and prompts). Note that this is consistent with the statistical results in Section~\ref{sec:t2i_results}. We show five examples in the figure. The first two images contain defective objects (paddle/bat), while the remaining images have unnatural object structures. 

\paragraph{Why Does SPA Improve Object Shapes?}
Interestingly, SPA can refine object shapes in addition to statistical image properties (e.g., lighting, color distribution). We hypothesize two possible reasons for this phenomenon. (A) By fixing the SNR at each timestep, diffusion models can better utilize signal components from earlier steps, resulting in refined object shapes. (B) In latent diffusion models, VAEs are known to have disentangled channel representations (colors, object shape, layout, lighting, etc.)~\cite{sdxl_latent}; therefore, aligning these channels can alter the overall content.

\section{Related Work}
\label{sec:related}

\subsection{Exposure Bias}

The concept of exposure bias was originally identified in autoregressive sequence generation, particularly in neural machine translation and language modeling~\cite{bengio2015scheduled, ranzato2016exposure}. In teacher forcing, models are trained to predict each token given ground-truth previous tokens, but at inference time, they must condition on their own predictions. This train-inference mismatch causes error accumulation, as models encounter inputs from a distribution different from that seen during training. 

Diffusion models exhibit a structurally identical problem. Ning \etal.~\cite{ning2024elucidating} explicitly identified this connection and proposed \emph{$\epsilon$-rescaling} to correct noise-level mismatch during inference. Li \etal~\cite{li2024error} analyzed error accumulation using an analytical model. Subsequent work has explored complementary corrections, as introduced in Section~\ref{sec:baselines}.

\subsection{Guidance-based Correction Methods}
Fundamentally, exposure bias stems from prediction errors in single-step predictions. Modifications to Classifier-Free Guidance (CFG)~\cite{ho2021cfg} aim to reduce such errors~\cite{wang2024analysis, chung2025cfgplus,kwon2025tcfg,sadat2025eliminating}. Although CFG is known as a truncation and error-correction method, it can also be a source of inference error, which often leads to oversaturation, unnatural contrast, and loss of fine details. 

Discriminator Guidance~\cite{kim2023discriminator} refines model scores using a specially trained discriminator. Although it is a powerful way to refine predictions during inference, it requires unstable adversarial training and incurs additional inference costs for the discriminator at every denoising step. In the context of training-free guidance, MPGD~\cite{he2024manifold} proposes to alleviate \emph{manifold error} using backpropagation through autoencoders, but this also requires additional inference costs. Note that our method is technically not mutually exclusive with these approaches, although empirical validation of their combination remains for future work.

\section{Conclusion}
\label{sec:conclusion}
We proposed Spectral Alignment (SPA) for diffusion models, which alleviates spectral mismatch and refines image quality during inference with minimal computational overhead.
We demonstrated the effectiveness of SPA across various models, from classic pixel-space models to modern flow-matching models. In addition, the concept of spectral mismatch that we introduced can serve as a lens for further improvements in model training, where inspection tools are valuable.

\paragraph{Limitations.} The current formulation of our method assumes a single target spectrum as a prior; however, the optimal spectral prior may vary depending on the target data distribution. For example, illustrations and medical images exhibit different frequency properties. For conditional generation, it may be possible to use different target spectra depending on the conditioning information. In this paper, we did not explore alternative guidance methods or loss weighting strategies. Although we adopted a simple DPS-based guidance approach, there is room to explore more sophisticated methods. Regarding loss weighting, one could design a weighting schedule for $\mathcal{L}_{spec}$ across frequency bands and timesteps.


\section{Acknowledgements}
We would like to thank Naoki Murata for carefully reading the manuscript and giving us valuable feedback.

%
%
\bibliographystyle{splncs04}
\bibliography{main}

\newpage
\appendix
\renewcommand{\thefigure}{A\arabic{figure}}
\setcounter{figure}{0} \setlength\textfloatsep{15pt}
\renewcommand{\thetable}{A\arabic{table}}
\setcounter{table}{0} \setlength\textfloatsep{15pt}

\title{[Appendix] Spectral Prior for Reducing Exposure Bias in Diffusion Models}

\author{}
\institute{}





\maketitle

\section{Details of the Regression of the Spectrum Model}
For the regression of our parametric spectrum model, we used \texttt{scipy.curve\_fit} in log space. The model is shown in Eq. (11) in the main document.
The average coefficients of determination ($R^2$ scores) across timesteps are shown in Table~\ref{tab:r2score}.

\begin{table}
\centering
\caption{$R^2\, \text{scores}$}
\begin{tabular}{c|cccccc}
    \toprule
    Models & DDPM & ADM & SD2.0 & SDXL & SD3.5 & FLUX.1 [dev] \\
    \midrule
    $R^2\,\text{scores}$ & 0.927 & 0.948 & 0.999 & 0.998 & 0.998 & 0.953 \\
    \bottomrule
\end{tabular}
\label{tab:r2score}
\end{table}

\begin{figure}[t]
    \begin{minipage}[b]{0.5\linewidth}
        \includegraphics[width=\linewidth]{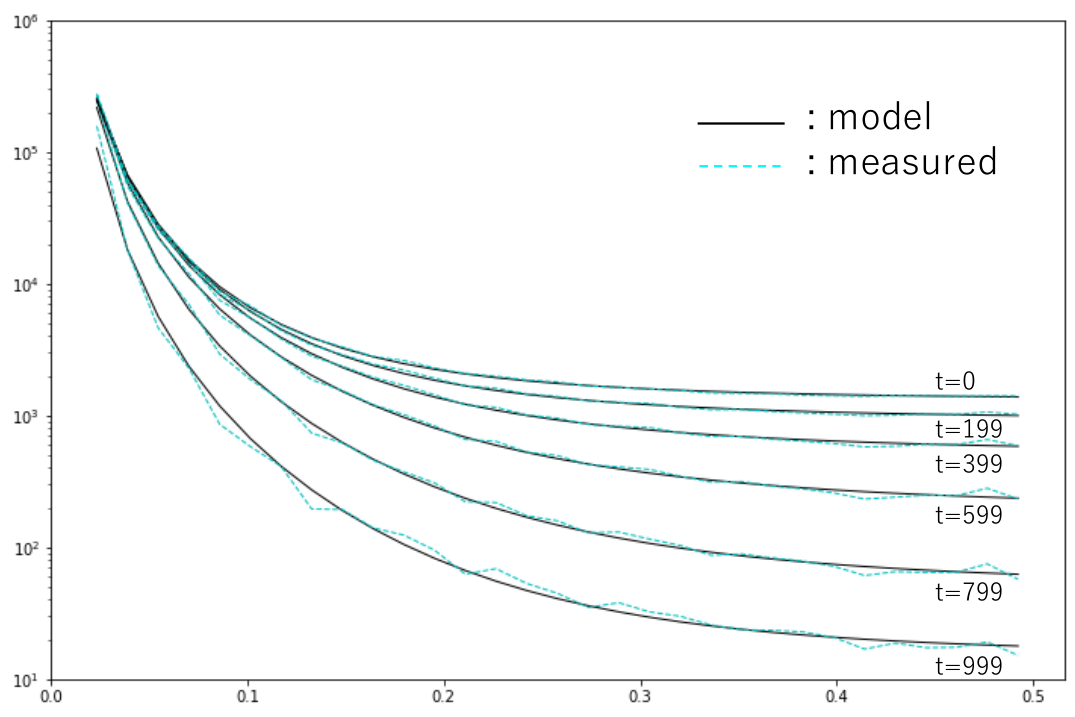}
        \caption{Fitting results with SD2.0.}
        \label{fig:fitting_sample}
    \end{minipage}
    \begin{minipage}[b]{0.5\linewidth}
        \includegraphics[width=\linewidth]{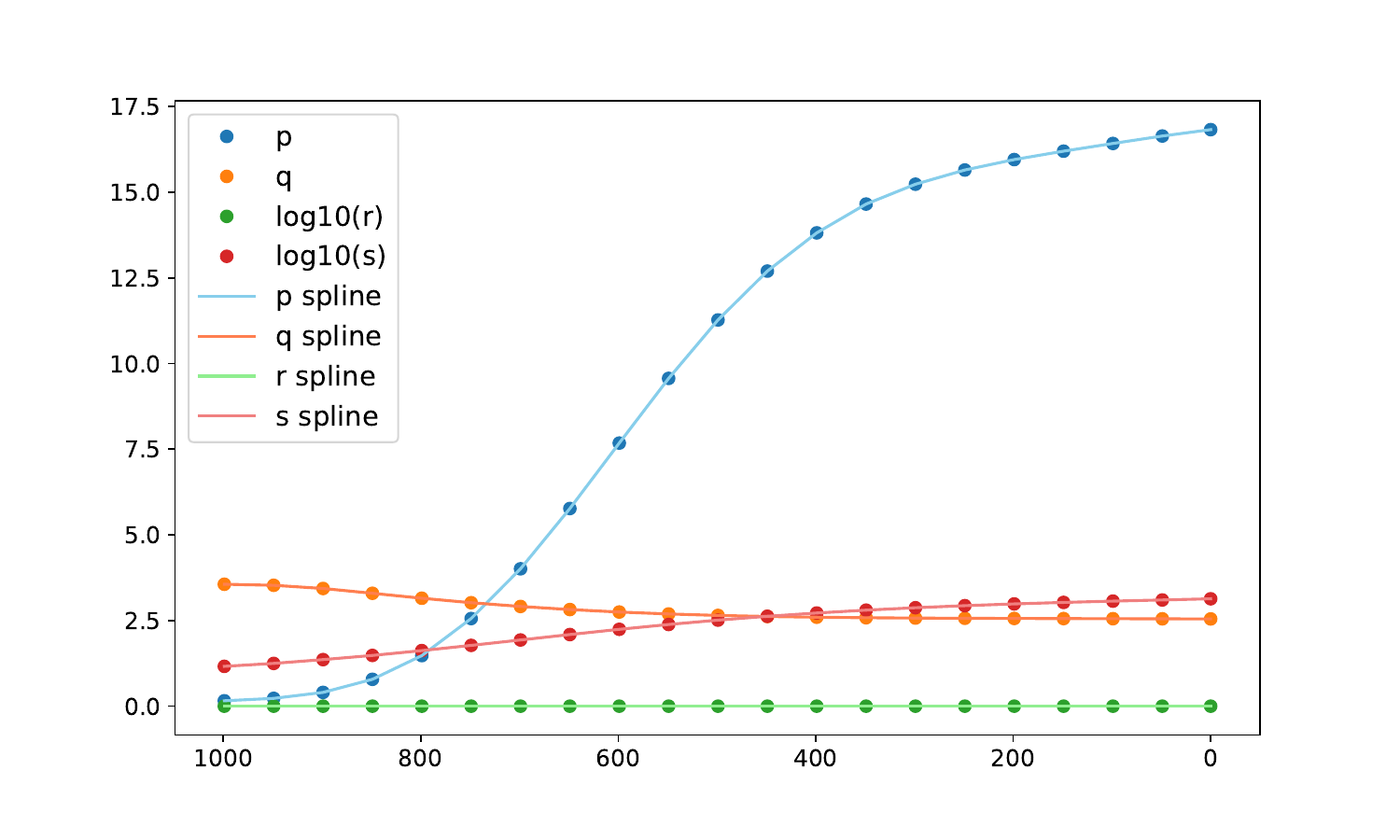}
        \caption{Interpolation results.}
        \label{fig:spline_sample}
    \end{minipage}
    \caption{Visualization of fitting results with SD2.0. Only a single channel is visualized.}
\end{figure}

Our spectrum model exhibits good fit ($R^2 > 0.9$) for all models, including latent diffusion models. Fig.~\ref{fig:fitting_sample} shows examples of fitting results on SD2.0.
As we mentioned in the main document (Section~4.1), the spectrum model also works well for $\hat{x}_{0\mid t}$ of latent diffusion models, as shown in this figure. Our spectrum model is based on the well-known property that the power spectrum of natural images follows a power law. This enables us to apply the same spectrum model to latent variables that share similar statistical properties to natural images.

Although we computed 10K samples for the regression, we still see fluctuations in the measured spectra. We obtain a smooth function by introducing the fitting step, which is expected to eliminate fluctuations and stabilize the following guidance phase. 

Fig.~\ref{fig:spline_sample} shows interpolation results of the parameters of the spectrum model: $p, q, r, s$. The parameters are smoothly interpolated over timesteps, which enables us to use an arbitrary number of denoising steps.


\section{Additional samples of Power Spectra During Inference}
\begin{figure}[ht]
\centering
    \begin{subfigure}[b]{\linewidth}
        \includegraphics[width=\linewidth]{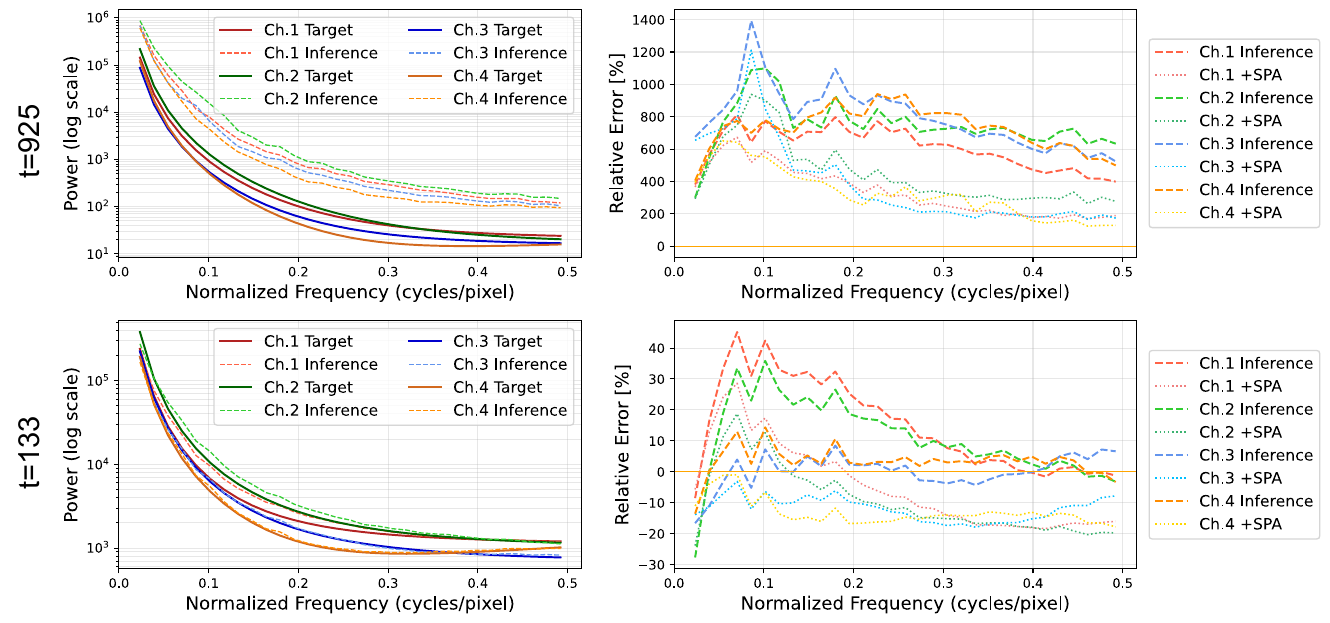}
        \caption{SD2.0}
    \end{subfigure}
    \begin{subfigure}[b]{\linewidth}
        \includegraphics[width=\linewidth]{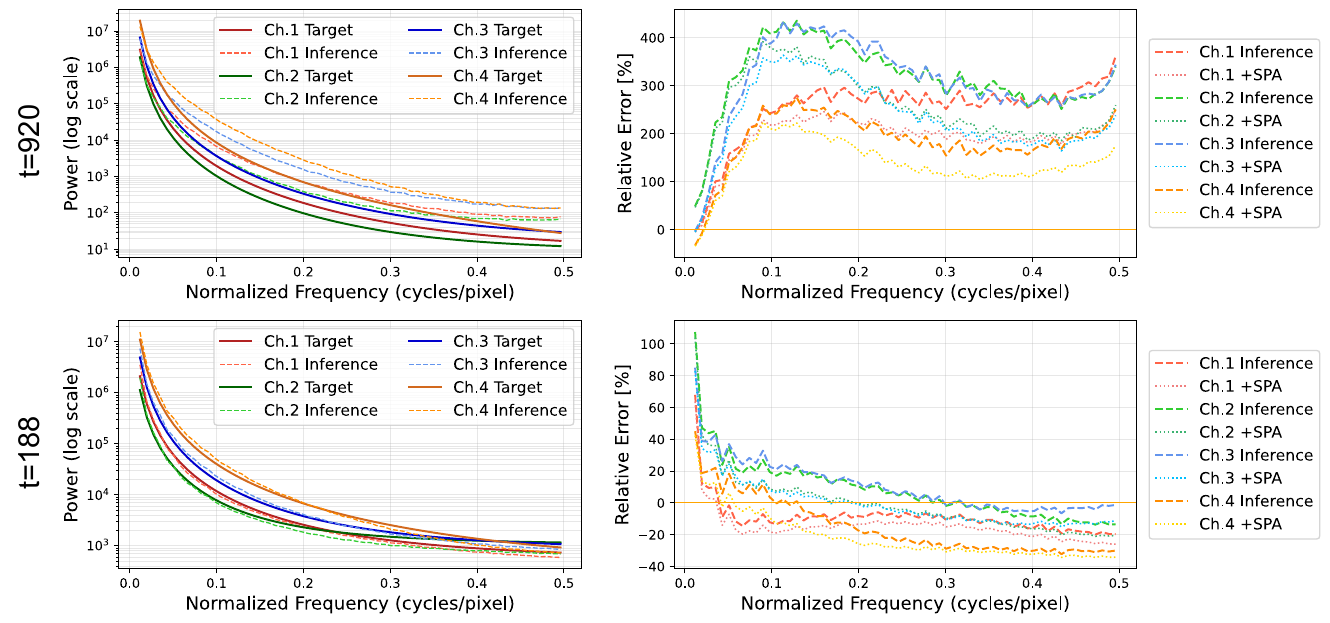}        
        \caption{SD3.5 (randomly selected four channels)}
    \end{subfigure}
    \caption{Additional examples of spectrum and relative error during inference. This corresponds to Figure~2 in the main document.}
    \label{fig:additional_spectrum}
\end{figure}

We show additional spectra from Stable Diffusion 2.0 and 3.5 in Figure~\ref{fig:additional_spectrum} (in addition to Figure~2 of the main document). The timesteps are not exactly the same between the models, because SD3.5 uses uneven timesteps by default. We used the closest timesteps instead. The errors from these models also behave differently from the other models. 

\newpage
\section{Analysis of the Hyperparameters}
In this section, we show the effect of changing the hyperparameters. Our method introduced two hyperparameters: the guidance strength $\eta$ and penalty intensity $a$ (see Algorithm 1 in the main document). 

\subsection{Analysis on the guidance strength $\eta$}
\begin{figure}[ht]
    \begin{subfigure}[b]{0.45\textwidth}
        \includegraphics[width=\linewidth]{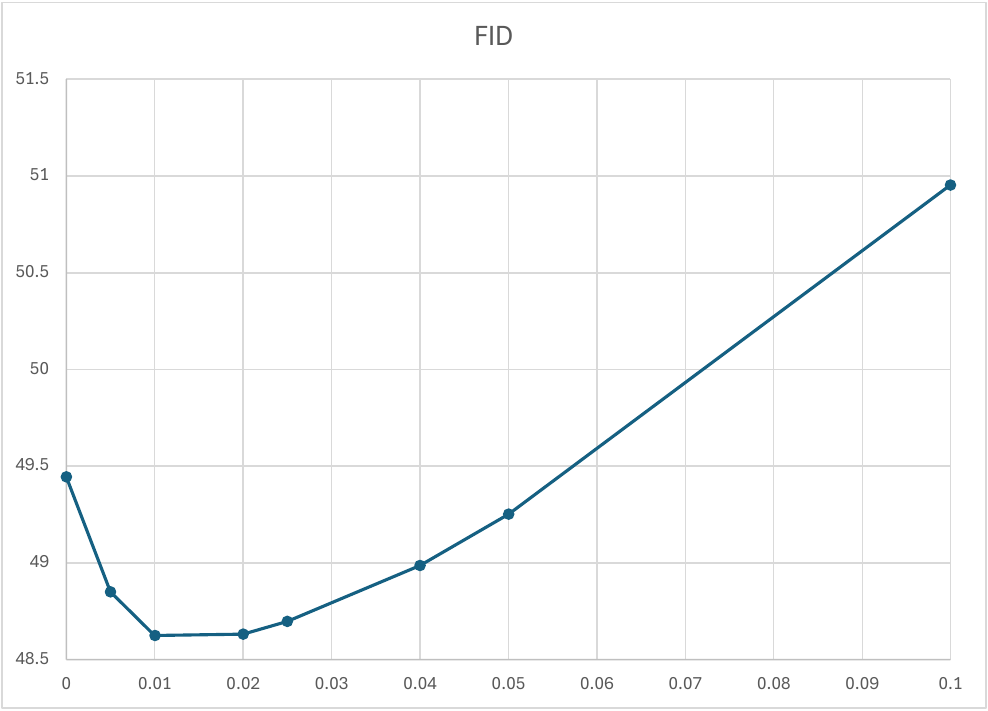}
    \end{subfigure}
    \hfill
    \begin{subfigure}[b]{0.45\textwidth}
        \includegraphics[width=\linewidth]{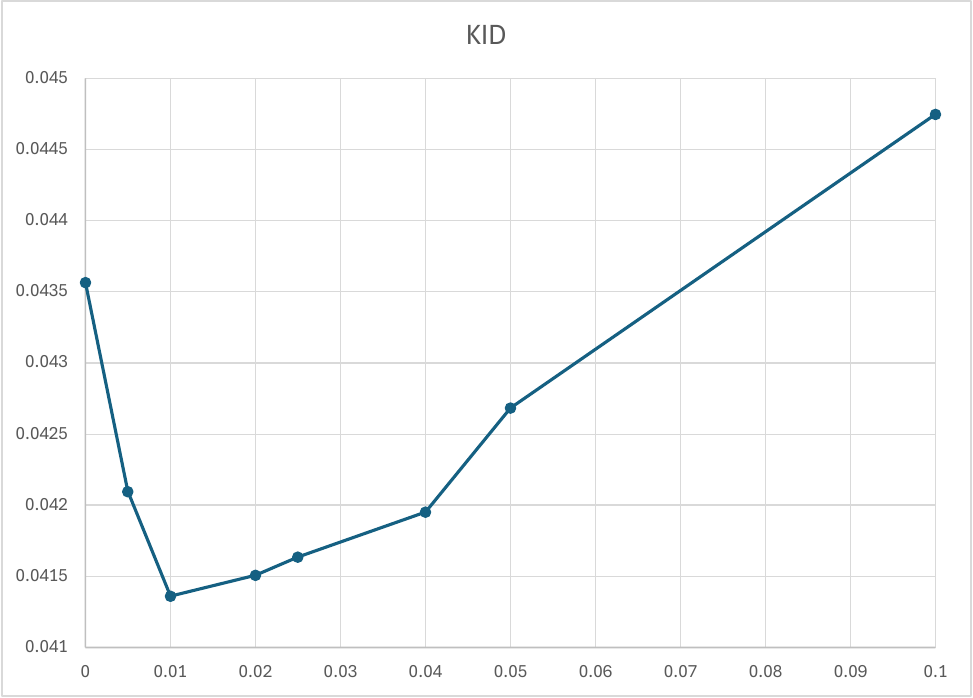}
    \end{subfigure}
    \caption{The effect of $\eta$ to FID and KID on DDPM.}
\label{fig:fid_vs_eta}
\end{figure}

We show the effect of guidance strength $\eta$ on FID and KID on DDPM in Figure~\ref{fig:fid_vs_eta}. It basically shows unimodality to the evaluation metrics, which makes hyperparameter selection straightforward. We set the penalty intensity $a=1$ (equivalent to ablating the penalty function, Equation~(10)) in this figure, since it is the optimal value for DDPM, and we can separate the effect of $\eta$ and $a$.

\subsection{Analysis on the penalty intensity $a$}
\begin{figure}[t]
\centering
\includegraphics[width=0.7\linewidth]{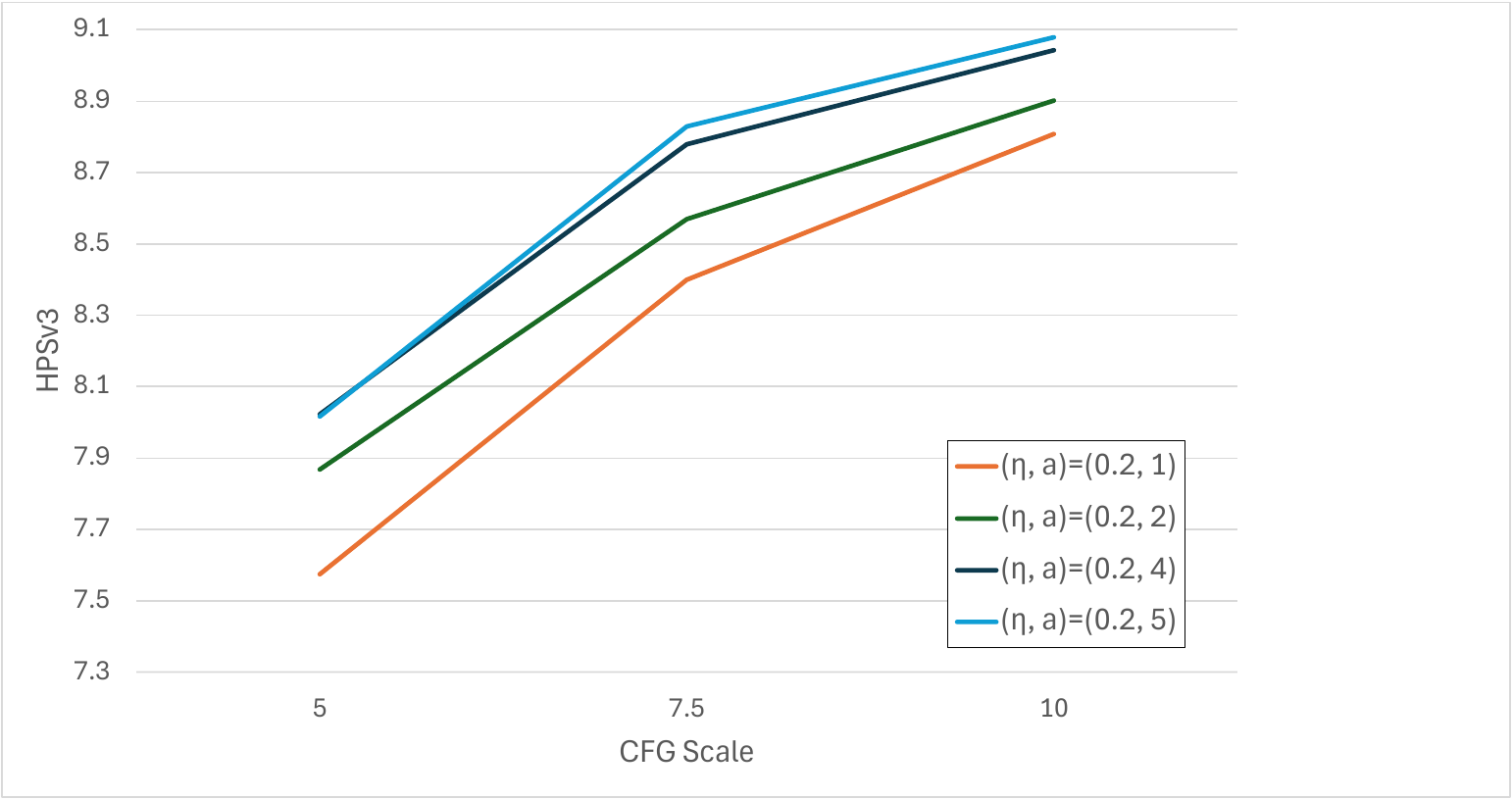}
\caption{Effect of penalty intensity $a$ and CFG scale to HPSv3 on SDXL.}
\label{fig:penalty_quant_sdxl}
\end{figure}

In this section, we analyze the effect of the penalty intensity $a$. We show the results with SDXL to also report the weak interaction between $a$ and the CFG scale.
We show the plot of HPSv3 vs $a$ in Figure~\ref{fig:penalty_quant_sdxl} and qualitative results in Figure~\ref{fig:penalty_qual_sdxl}. We kept $\eta=0.2$, which is the optimal value for SDXL. Note that, unlike the previous section, we can not set $\eta=0$, since the penalty function is part of the guidance.

\paragraph{Quantitative Results}
Figure~\ref{fig:penalty_quant_sdxl} shows quantitative results of HPSv3 with different $a$ values and CFG scales. The penalty intensity $a$ enhances the score, as shown in this figure. The improvement in the scores gradually decreases as the parameter $a$ increases. 

\begin{figure}[ht]
\centering
\includegraphics[width=\linewidth]{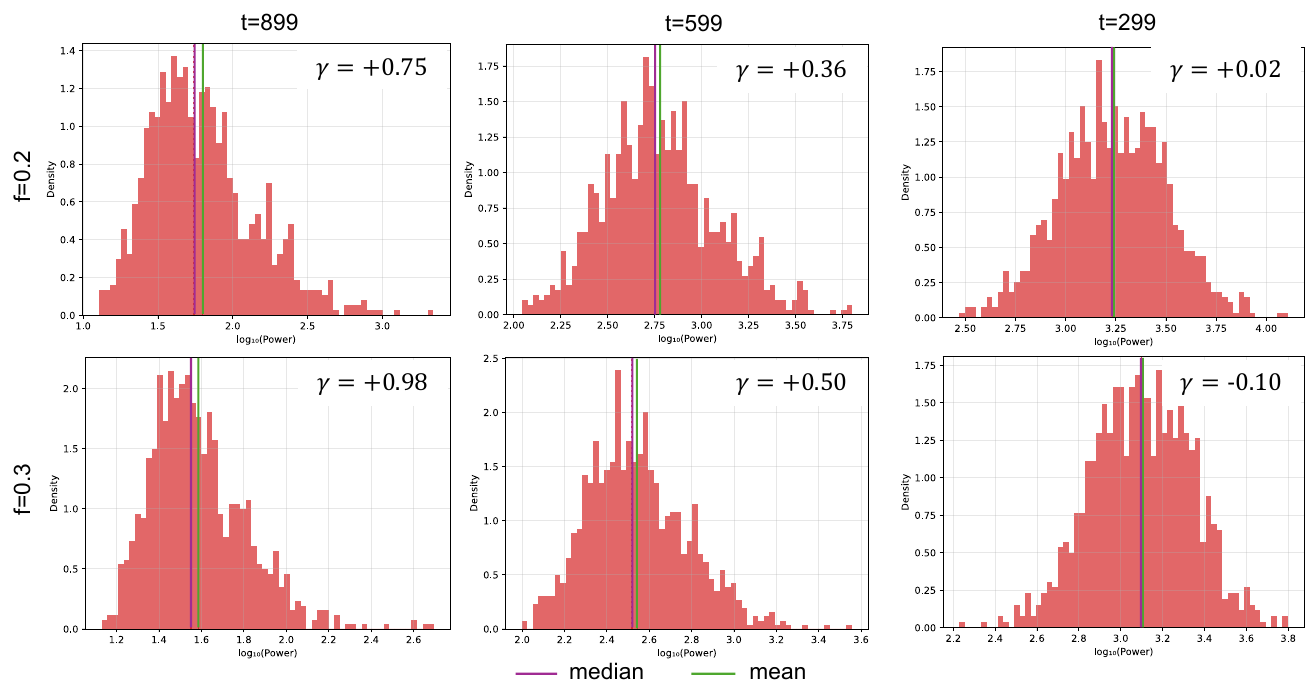}
\caption{Histograms of representative frequencies ($f=0.2, 0.3$) and timesteps ($t=899, 599, 299$). These plots are from one of the channels of SDXL. They typically show positive skewness, especially in large timesteps. The purple line indicates the median, and the green line indicates the mean of the data. $\gamma$ is the skewness of each histogram.}
\label{fig:skewness}
\end{figure}

\begin{figure}[ht]
\centering
\includegraphics[width=1.0\linewidth]{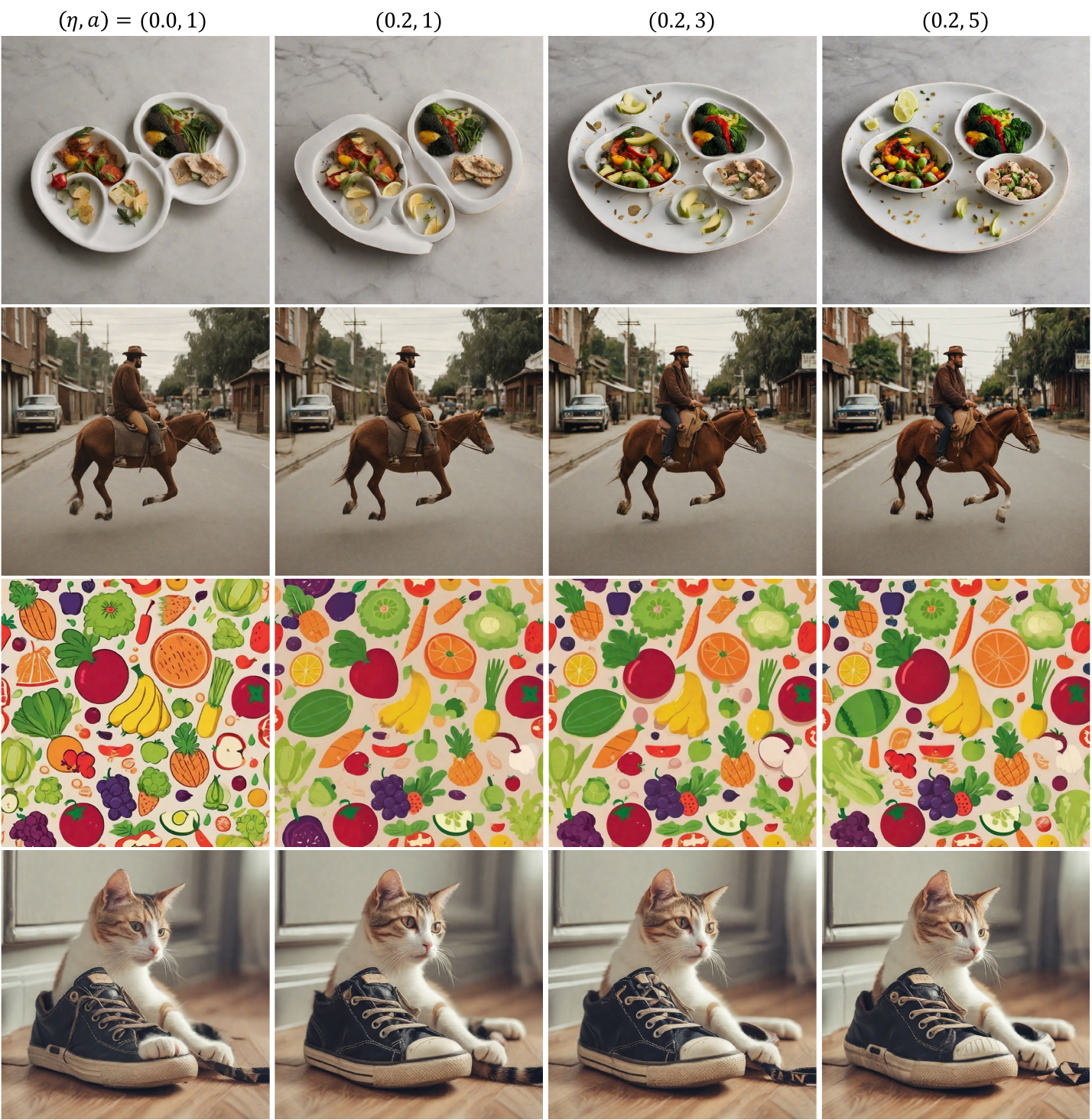}
\caption{Qualitative result with different penalty intensity $a$ on SDXL. CFG scale $w=7.5$ is used. Prompts: (1) ``\textit{a couple of plates that have some food on them}'', (2) ``\textit{A man riding on the back of a brown horse down a street}'', (3) ``\textit{Fruits and vegetables are on the cutting board}'', and (4) ``\textit{A cat is laying on the floor with its paws in a shoe}''.}
\label{fig:penalty_qual_sdxl}
\end{figure}

\paragraph{Skewness of a Spectrum}
As we discussed in Section~3.3 of the main document, we designed our asymmetric penalty function due to the skewness of the measured spectra. We show examples of the histograms in Figure~\ref{fig:skewness}. Most of the measured spectra show positive skewness and ``$\text{mean} > \text{median}$'', especially in $t > 300$. As we see in this figure, skewness varies across timesteps and frequency bands. Therefore, there is room for changing the penalty intensity $a$ depending on timesteps and frequency bands, which we leave for future work.
Furthermore, estimating the optimal $a$ from skewness could also be an interesting future direction.

\paragraph{Qualitative Results}
Figure~\ref{fig:penalty_qual_sdxl} shows qualitative results with different penalty intensity $a$ on SDXL. Larger $a$ (\ie, stronger penalty) leads to better object shapes, more pronounced colors, and greater sharpness/contrast.

\newpage
\section{Implementation Details of the Baseline Methods}
We adapted the three baseline methods for latent diffusion models with minimal modification. Please refer to Section~4.1 baselines in the main document. For $\epsilon$-rescaling, we applied the scaling to the model prediction after CFG. For time-shift sampler and wavelet regulation, we applied each method to the latent variable before the sampler step ($\text{Denoise}(x_t, \epsilon^{\text{CFG}}, t)$ in  line~9, Algorithm~1).


\section{Additional Qualitative Results}
\begin{figure}[htbp]
  \centering

  \begin{subfigure}{\linewidth}
    \centering
    \includegraphics[width=\linewidth]{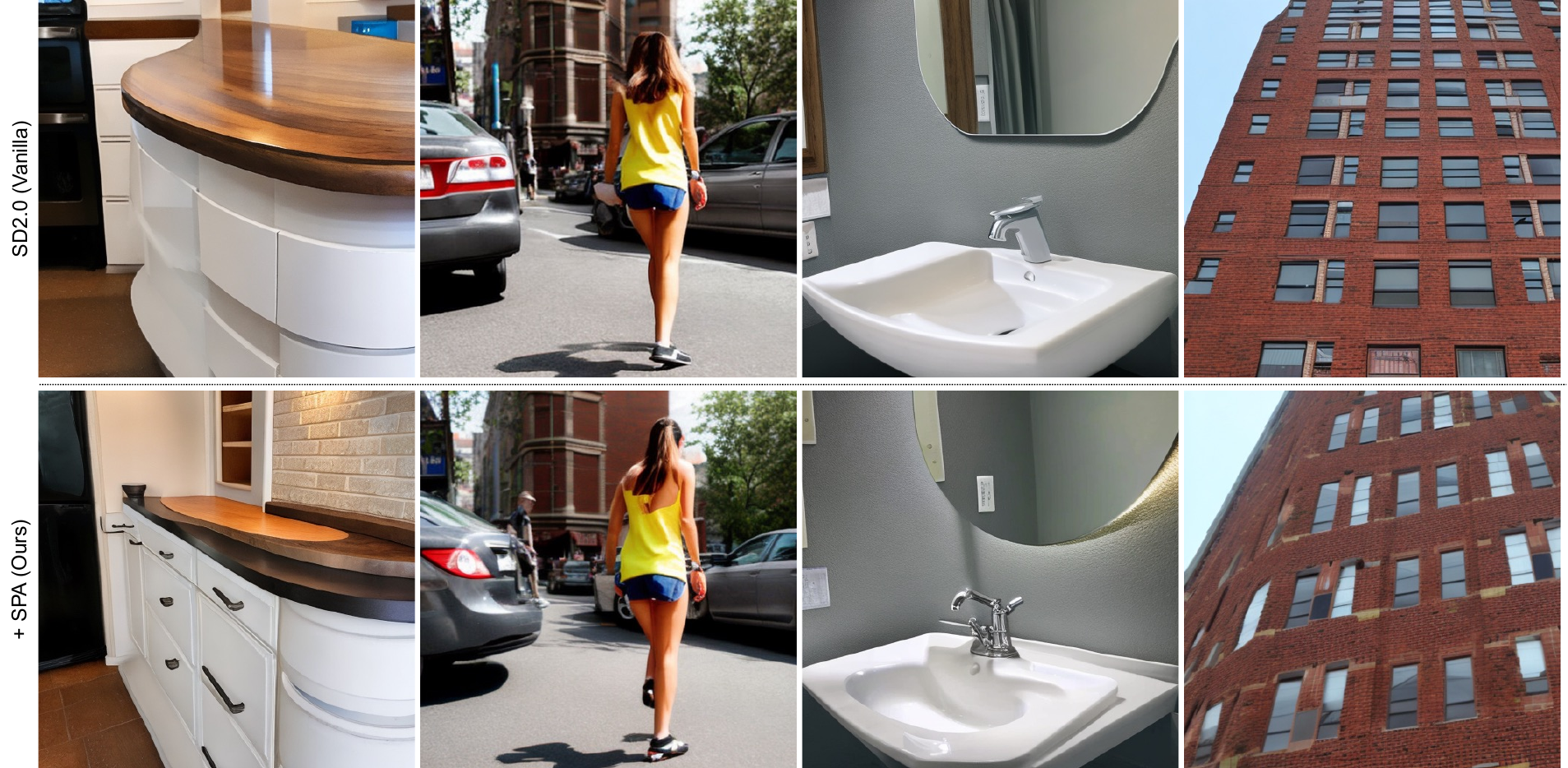}
    \caption{Samples from SD2.0.}
    \label{fig:sd20}
  \end{subfigure}

  \vspace{1em} 

  \begin{subfigure}{\linewidth}
    \centering
    \includegraphics[width=\linewidth]{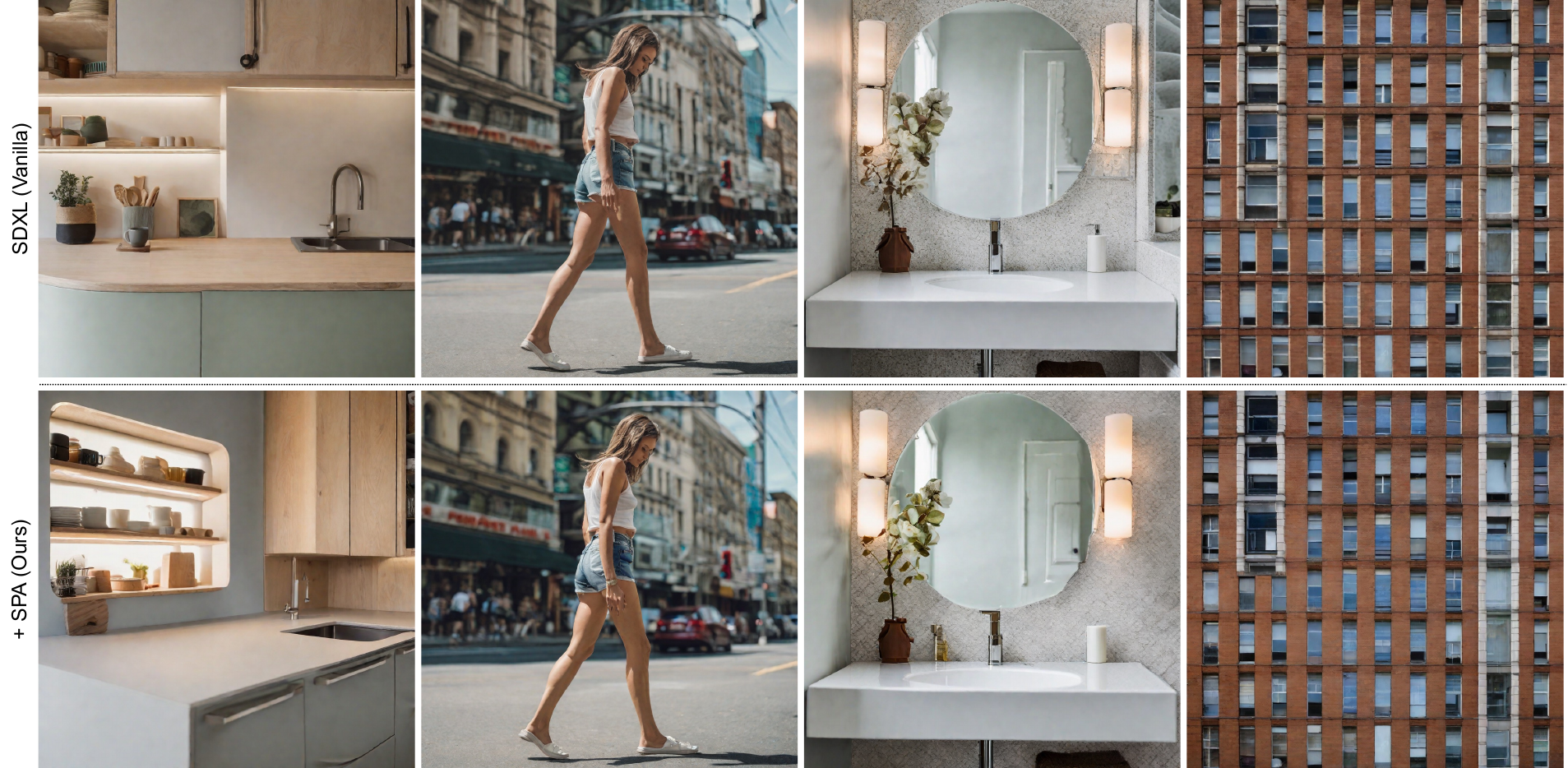}
    \caption{Samples from SDXL.}
    \label{fig:sdxl}
  \end{subfigure}

  \begin{subfigure}{\linewidth}
    \centering
    \includegraphics[width=\linewidth]{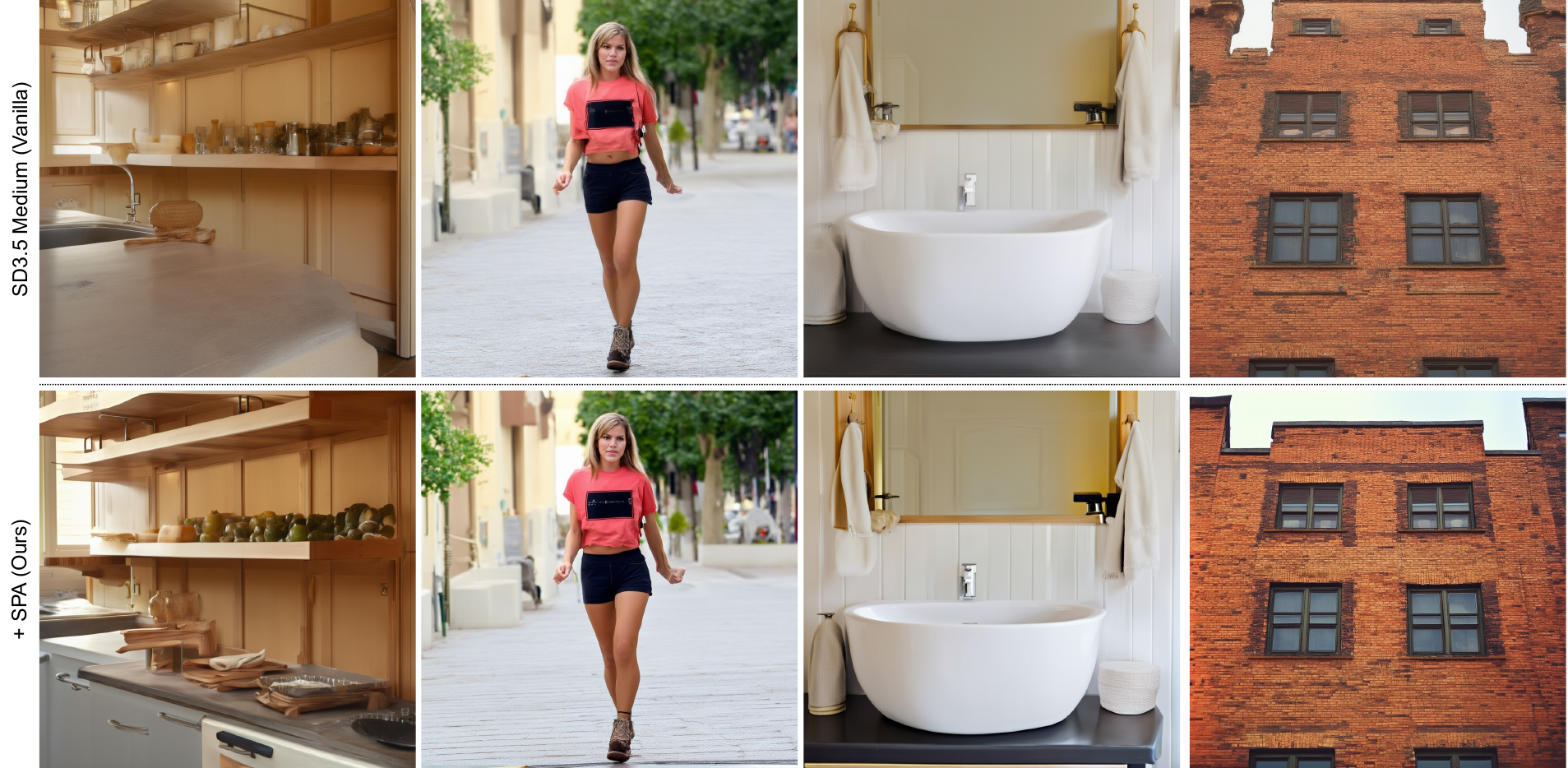}
    \caption{Samples from SD3.5.}
    \label{fig:sd3.5}
  \end{subfigure}

  \caption{Additional qualitative samples from text-to-image models.}
  \label{fig:t2i_additional_samples}
\end{figure}

\begin{figure}[htbp]
  \centering
  \begin{subfigure}{\linewidth}
    \centering
    \includegraphics[width=\linewidth]{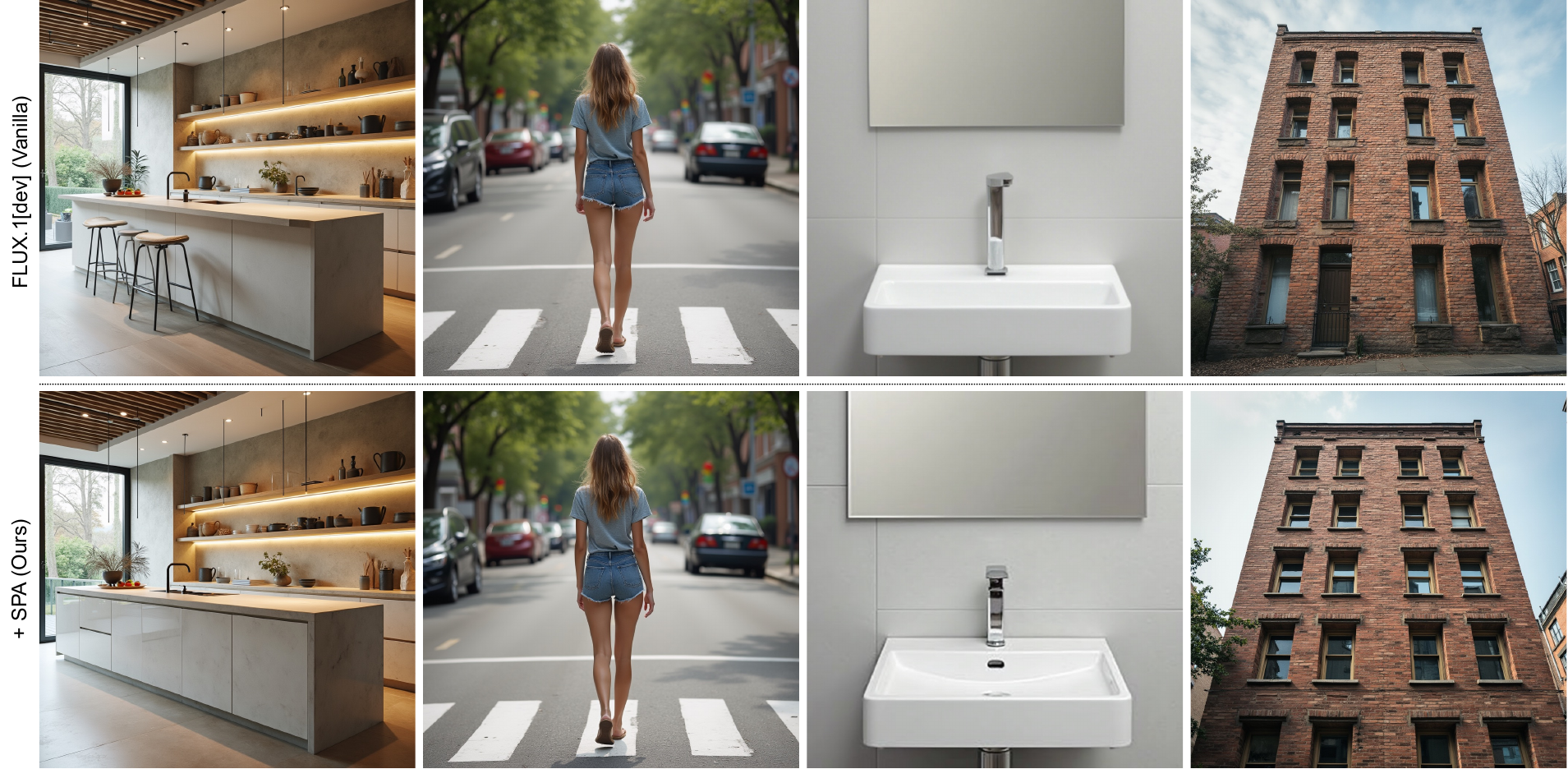}
    \caption{Samples from FLUX.1[dev] (distilled CFG).}
    \label{fig:flux_distilled}
  \end{subfigure}

  \begin{subfigure}{\linewidth}
    \centering
    \includegraphics[width=\linewidth]{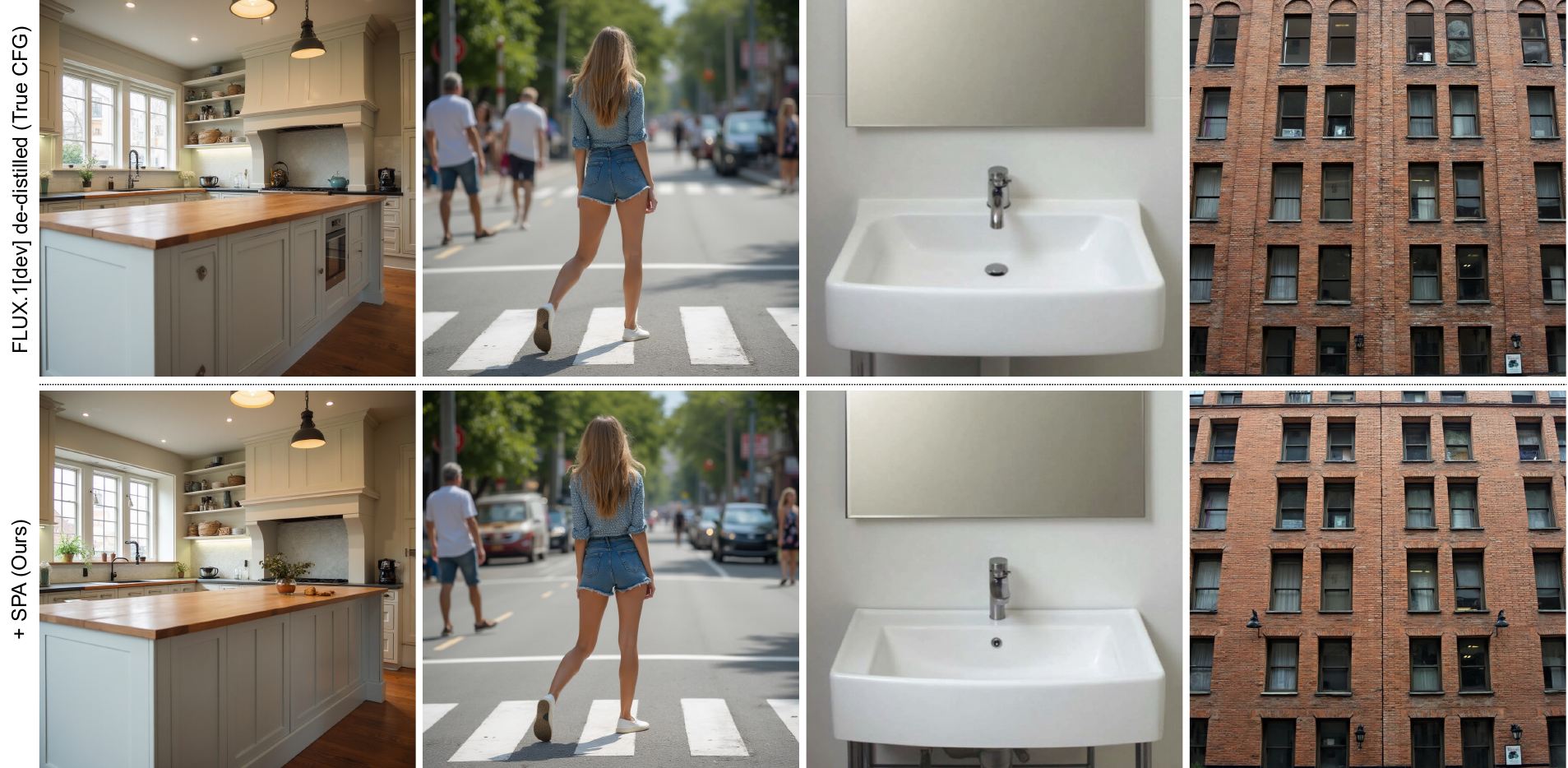}
    \caption{Samples from FLUX.1[dev] ``de-distilled'' version, which re-introduced standard CFG by finetuning FLUX.1[dev]}
    \label{fig:flux_dedistill}
  \end{subfigure}

  \caption{Additional qualitative samples from FLUX models.}
  \label{fig:flux_comparison}
\end{figure}

\begin{figure}[ht]
\centering
\includegraphics[width=\linewidth]{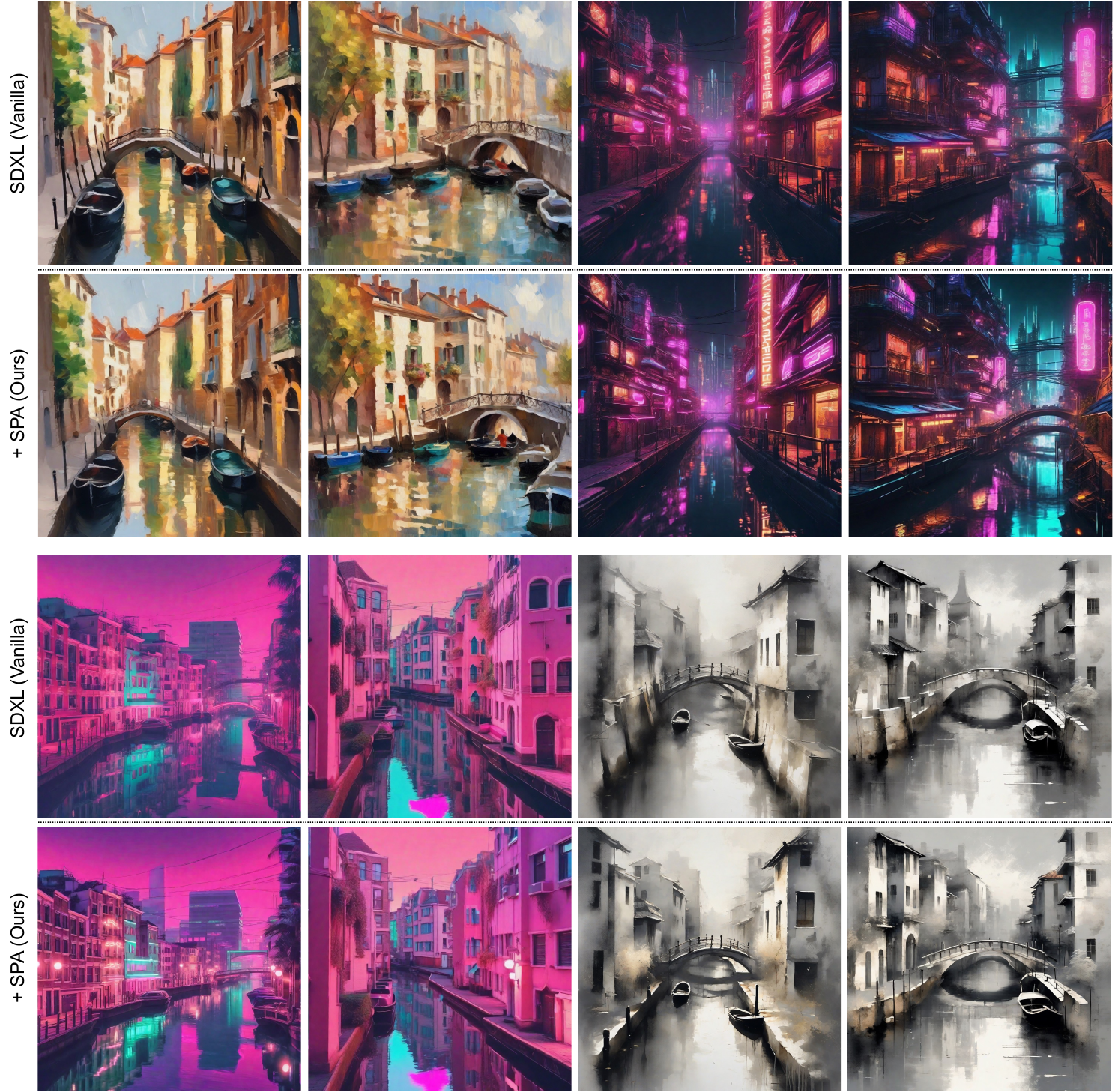}
\caption{Styles other than photorealistic images with SDXL. SPA has no negative effects on styles other than photorealistic images. We use a base prompt followed by a style prompt. \\Base prompt: beautiful city with canals\\ Style prompts: (1) ``\textit{Impressionist painting, loose brushstrokes, light-focused color palette}'', (2) ``\textit{Cyberpunk aesthetic, neon lights, urban night, futuristic cityscape}'', (3) ``\textit{Vaporwave aesthetic, pastel neon, retro internet nostalgia}'', and (4) ``\textit{elegant brush strokes, delicate gray tones, large white space, washi paper texture, Zen}''  (numbering in a z-shaped)}
\label{fig:user_prompts}
\end{figure}

Figure~\ref{fig:t2i_additional_samples}, \ref{fig:flux_comparison}, and \ref{fig:user_prompts} show additional qualitative samples from text-to-image models. Please zoom in for details. We use four prompts from the MS COCO dataset for all the models. (1) ``\textit{A kitchen counter with a rounded edge and shelves}'' (2) ``\textit{A woman walking across a street in short shorts}'', (3) ``\textit{A white bathroom sink sitting under a mirror}'', and (4) ``\textit{A very tall brick building with bricked up windows}''.

SPA enhances the original images in most samples and rarely affects them negatively. 

Figure~\ref{fig:flux_dedistill} shows results using the ``dedistilled'' checkpoint of FLUX.1[dev] (\href{https://huggingface.co/nyanko7/flux-dev-de-distill}{nyanko7/flux-dev-de-distill}). This checkpoint re-introduced the standard two-pass CFG by deleting the weight embedding used for guidance distillation and is finetuned with some portion of blank conditioning. The difference made by SPA tends to be larger with this checkpoint than the official guidance-distilled version. 
Figure~\ref{fig:user_prompts} shows results with styles other than the standard photo-real style. Although SPA guides samples toward the single target spectrum (\emph{spectrum prior}), it does not negatively affect sampling of artistic styles. 



%
%

\end{document}